%% file: main.tex
\documentclass[]{fairmeta}

\input{math_commands.tex}

\usepackage{url}
\usepackage{graphicx}

\usepackage{tikz}
\usepackage{hyperref}

\usepackage{subcaption}

\usepackage{enumitem}

\newcommand*\circled[1]{\tikz[baseline=(char.base)]{\node[shape=circle,draw,inner sep=0.8pt] (char) {#1};}}

\usepackage[nameinlink]{cleveref}

\input{macros}

\title{MEXMA: Token-level objectives improve sentence representations}

\author[1,2]{João Maria Janeiro}
\author[2]{Benjamin Piwowarski}
\author[2,3]{Patrick Gallinari}
\author[1]{Loïc Barrault}
\affiliation[1]{Meta AI}
\affiliation[2]{Sorbonne Université, CNRS, ISIR, F-75005 Paris, France}
\affiliation[3]{Criteo AI Lab, Paris, France}

\abstract{\input{sections/abstract}}

\correspondence{João Maria Janeiro at \email{joaojaneiro@meta.com}}

\usepackage[nameinlink]{cleveref}

\begin{document}

\maketitle

\input{sections/introduction}
\input{sections/related_work}
\input{sections/method}
\input{sections/evaluation}
\input{sections/experiments}
\input{sections/conclusion}

\input{sections/acknowledgments}

\clearpage
\newpage
\bibliographystyle{assets/plainnat}
\bibliography{paper.bib}

\clearpage
\newpage
\beginappendix

\input{sections/appendix_experimental_setup}
\clearpage
\newpage
\input{sections/appendix_ablations}

\clearpage
\newpage
\input{sections/appendix_language_info}

\clearpage
\newpage
\input{sections/appendix_datasets}

\clearpage
\newpage
\input{sections/appendix_mteb_datasets}
\clearpage
\newpage
\input{sections/appendix_token_level_analysis}
\clearpage
\newpage
\input{sections/appendix_sentence_analysis}
\clearpage
\newpage
\input{sections/appendix_other_architectures}

\end{document}

%% file: math_commands.tex
\usepackage{amsmath,amsfonts,bm}

\def\eqref#1{equation~\ref{#1}}

\def\1{\bm{1}}

\DeclareMathAlphabet{\mathsfit}{\encodingdefault}{\sfdefault}{m}{sl}
\SetMathAlphabet{\mathsfit}{bold}{\encodingdefault}{\sfdefault}{bx}{n}

%% file: macros.tex
\newcommand*\langA{language $\mathcal{A}$}
\newcommand*\langB{language $\mathcal{B}$}
\newcommand*\sentA{$S_\mathcal{A}$}
\newcommand*\sentB{$S_\mathcal{B}$}

\newcommand*\bd[1]{\textbf{#1}}

%% file: sections/abstract.tex
Current pre-trained cross-lingual sentence encoders approaches use sentence-level objectives only.
This can lead to loss of information, especially for tokens, which then degrades the sentence representation.
We propose MEXMA, a novel approach that integrates both sentence-level and token-level objectives.
The sentence representation in one language is used to predict masked tokens in another language, with both the sentence representation and \textit{all tokens directly updating the encoder}.
We show that adding token-level objectives greatly improves the sentence representation quality across several tasks.
Our approach outperforms current pre-trained cross-lingual sentence encoders on bi-text mining as well as several downstream tasks.
We also analyse the information encoded in our tokens, and how the sentence representation is built from them.

%% file: sections/introduction.tex
\section{Introduction}
Creating general-purpose multilingual embeddings has attracted significant attention from the research community in recent years, driven by the growing need for efficient and effective cross-lingual representations.
Cross-Lingual Sentence Encoders (CLSE) create fixed-size sentence representations that are able to capture the relevant information in a sentence, and are aligned across languages.
By capturing relevant sentence information in a shared multilingual space, these aligned representations enable \textit{efficient} comparison and retrieval based on distance measures, thereby facilitating their effective utilization in various downstream applications.

Current CLSE \citep{duquenne2023sonar, feng-etal-2022-language} typically build upon pre-trained encoders, often language models \citep{conneau-etal-2020-unsupervised, devlin-etal-2019-bert} or translation models \citep{nllbteam2022languageleftbehindscaling}.
These pre-trained encoders have been trained using objectives that focus on individual words or tokens, i.e. token-level objectives.
Examples of such objectives include unmasking, where the model is required to predict each token individually, and \textit{all predictions} are used to \textit{update the encoder} directly.
However, \citet{muennighoff-etal-2023-mteb, pmlr-v119-hu20b} show that pre-trained encoders without objectives that consider entire sentences, i.e. sentence-level objectives, do not create good sentence representations.
This means that CLSE need to train sentence-level representations, in order to effectively capture the relevant information of the sentences.

Although CLSE start from encoders pre-trained with token-level objectives, they are commonly trained with sentence-level objectives that \textit{only update the encoder through the sentence representation} \citep{duquenne2023sonar, feng-etal-2022-language, yang2019improvingmultilingualsentenceembedding, LASER}, without any objective for each token individually.
We hypothesize that token-level objectives should be kept during the training of CLSE, coupled with the sentence-level objectives, to better update the encoder and improve sentence representation quality and alignment.
The intuition is that only using sentence-level objectives leads to a degradation of token level information, especially lexical information, which in turn can impact the sentence representation.

Recently, there have been approaches exploring the use of both token-level and sentence-level objectives for better sentence representations.
DAP \citep{dap-li-etal-2023-dual} uses both objectives, but the token-level objective is only used to update the token representations in the encoder, without influencing directly the sentence representation.
On the other hand, RetroMAE \citep{xiao-etal-2022-retromae} also employs both objectives, but uses two different token objectives to update the individual tokens and the sentence, with the latter having to be created from a masked input.

To effectively combine token and sentence-level objectives, we propose MEXMA, a new approach that uses the sentence representation in one language to predict masked tokens in another language, and uses both the sentence and tokens' information to update the encoder.
This token-level objective is combined with a sentence-level objective to enforce sentence alignment across languages.

Our approach outperforms state-of-the-art pre-trained cross-lingual sentence encoders, LaBSE and SONAR on several key tasks, including bitext mining, classification, and pair classification.
Specifically, we report notable gains on the xsim++ benchmark computed over the FLORES200 test set, where MEXMA achieves an error rate of 9.60\%, surpassing SONAR's 12.08\%.
Additionally, in classification tasks evaluated on MTEB and SentEval, MEXMA achieves an accuracy of 65.35\% compared to SONAR's 63.02\%.
The larger supervision in MEXMA enables training smaller models with better alignment than LaBSE ($\approx$2$\times$ size) and close to SONAR's performance ($\approx$3$\times$ size).

Our main contributions are:
\begin{itemize}[topsep=0pt,itemsep=2pt,parsep=2pt]
    \item We introduce a novel architecture leveraging both sentence-level and token-level objectives outperforming current approaches.
    \item We perform ablation studies that show the impact of token-level objectives on the sentence-level representations performance.
    \item We provide an extensive analysis of the inner working of our model, by analysing its tokens' contents, and the way the sentence embedding is built.
    We show that as a byproduct of our training, individual tokens are also well aligned across languages.
    \item We show that our approach can also be coupled with existing alignment approaches, specifically contrastive learning, and improve its quality.
\end{itemize}

%% file: sections/related_work.tex
\section{Related Work}

Sentence embeddings have been well studied in the last decade.
Initially, recurrent networks were trained to predict previous and next sentence \citep{kiros2015skipthoughtvectors} or sentence entailment \citep{conneau2017infersent}.
Universal Sentence Encoder \citep{cer2018unisentenc} trains a transformer network on both tasks. 
\citet{sentenceBERT} propose to continue the training of a BERT model to include a sentence-level objective.
These initial works have been extended to multilingual settings, to capture the relevant information in the sentences, and to have aligned representations across languages.
These new approaches are called cross-lingual sentence encoder.
We describe those works next.

\paragraph{\textsc{Update via sentence representation}}
Most current cross-lingual sentence encoder approaches only update their encoder via the sentence representation objective, without having any token-level objective in the output of the encoder that would update each token individually \citep{guo-etal-2018-effective, yang2019improvingmultilingualsentenceembedding, feng-etal-2022-language, LASER, duquenne2023sonar, heffernan-etal-2022-bitext}.
They are most commonly based on contrastive learning \citep{contrastive_learning} methods, that aim to reduce the distance between positive pairs (translations) and increase the distance between negative pairs (non-translations) \citep{guo-etal-2018-effective, yang2019improvingmultilingualsentenceembedding, feng-etal-2022-language}.
Notably, LaBSE \citep{feng-etal-2022-language} uses the contrastive loss, with the additive margin softmax approach of \citet{yang2019improvingmultilingualsentenceembedding}.
A common non-contrastive solution is to use translation \citep{LASER, duquenne2023sonar} with a fixed-size sentence representation after the encoder (bottleneck), assuming that a model can translate a sentence into many languages only if a good sentence-level conceptual representation is learned.
The bottleneck however prevents gradients from the decoder to directly update the individual token representations of the encoder, which we hypothesize leads to a degradation of token level information and consequently of the sentence representation.
Our method also uses a sentence representation as context for the unmasking, but allows direct token-level gradients to propagate to the encoder token representations.

\paragraph{\textsc{Update via sentence and token representations}}
Recent approaches \citep{dap-li-etal-2023-dual, xiao-etal-2022-retromae} have shown that combining token and sentence level objectives can improve sentence representations.
RetroMAE \citep{xiao-etal-2022-retromae}, is an Information Retrieval (IR) method that utilizes fixed-size sentence representations to guide token unmasking, demonstrating its effectiveness in enhancing sentence representation quality.
The encoder itself is only updated by its own MLM loss with light masking (forcing the sentence representation to come from a masked input) and the sentence representation, but not from the direct token-level gradients of the heavy unmasking with the sentence representation as context.
DAP \citep{dap-li-etal-2023-dual} proposes to jointly align tokens and sentence representations.
It performs unmasking with all tokens of the other language as context, which means it updates the encoder with each token individually, however, it relies exclusively on the contrastive loss to update the sentence representations, and the sentence representation is not used to perform the token unmasking.
In our work, we show that sentence and token-level objectives can be much more intertwined, with both individual tokens and the sentence representation updating the encoder, and each other.

%% file: sections/method.tex
\section{Methodology} \label{section:methodology}
\begin{figure*}
    \centering
    \includegraphics[width=0.9\textwidth]{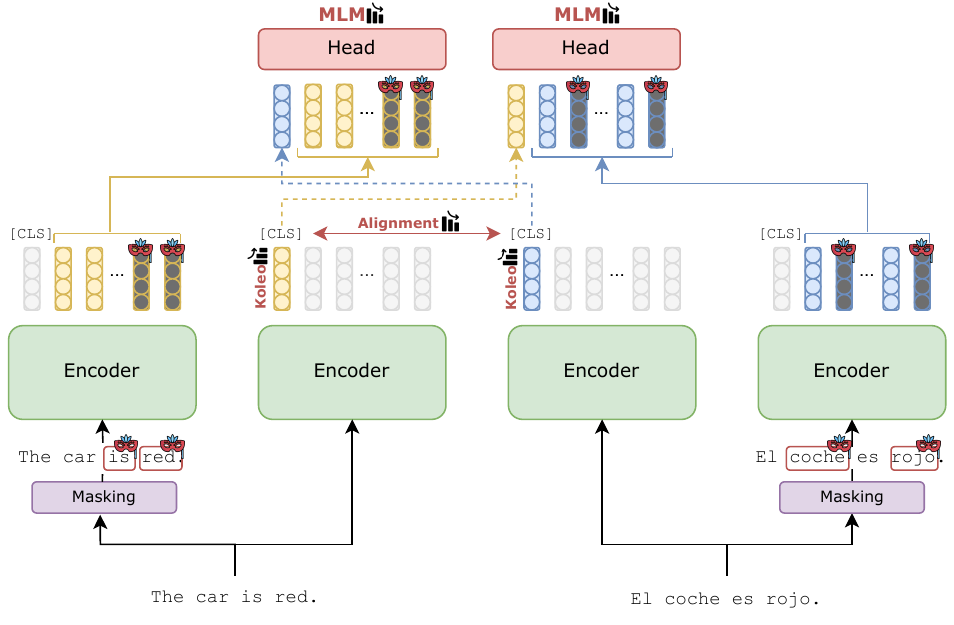}
    \caption{MEXMA architecture. Given two translations, we create two views for each, a masked and a clean version (symmetrical architecture), and use the sentence representations from one language to unmask the other (cross-unmasking). We align the clean sentence representations via the alignment loss, and increase the usage of the space with the KoLeo loss.}
    \label{fig:methodology/sym_mexma_architecture}
\end{figure*}
We propose MEXMA, a novel multilingual alignment technique based on both token-level and sentence-level objectives.
The goal is to create a sentence representation that is able to encode the syntactic, semantic and lexical information in a sentence, with representations well aligned across languages.
To achieve this goal, inspired by monolingual masked auto-encoding techniques~\citep{xiao-etal-2022-retromae}, we use the sentence representation in one language to unmask the tokens in another language, updating both the sentence and individual tokens, while forcing the sentence representation to encode the relevant parts of the sentence.
Using masking also allows us to use a non-contrastive loss to align sentence representations, since it prevents the collapse.
Both sentence and token-level objectives are used to improve the quality of the sentence representation.
Our architecture is depicted in \Cref{fig:methodology/sym_mexma_architecture}, and is composed of several components, that we describe now.
For the explanation, we refer to inputs (and the output of their encoders) that have no masking as \textit{clean}, and \textit{masked} for their masked counterparts.
Additionally, we consider two languages, \langA{} and \langB{}, which are associated with the sentence representations \sentA{} and \sentB{} (from the clean encoders).

\paragraph{\textsc{The cross-unmasking}}
To ensure that our sentence vector captures the meaningful information of the sentence, we mask a significant portion of the input tokens in \langA{}.
This makes it challenging for the encoder and the MLM head to recover the missing tokens without any additional context.
To overcome this challenge, we provide the unmasking head with the sentence vector \sentB{}, derived from the clean sentence in \langB{}.
This forces the model to leverage the information in \sentB{} to predict the masked tokens in \langA{}.
By doing so, we encourage the sentence vector to capture the essential information of the sentence.
Furthermore, by alternating languages, we enforce the sentence vector to encode information that is useful across languages.
We formulate this component into a symmetrical cross-entropy loss (CE), applied over the outputs of the encoders:
\begin{equation*}
    \mathcal{L}_{mlm} = CE([S_{B}, \hat{A}], A) + CE([S_{A}, \hat{B}], B) ,
\end{equation*}
where $\hat{A}$ and $\hat{B}$ are the outputs of the masked encoders without the CLS embedding, A and B the token targets, and $[X,Y]$ represents the concatenation of X and Y.

\paragraph{\textsc{The alignment loss}}
The cross-unmasking generates an implicit alignment due to the switching of languages to perform the unmasking.
However, as is, that implicit alignment does not strongly enforce the same sentence representations in two different languages to be equal in the embedding space.
Following SONAR~\cite{duquenne2023sonar}, to further reinforce the spatial proximity of semantically equivalent sentences across languages, we use an additional non-contrastive alignment objective.
The two losses, unmasking and alignment, complement each other to provide both aligned and meaningful vector representations of sentences in multiple languages.
We formulate this component as a Mean Squared Error (MSE) loss between sentence representations:
\begin{equation*}
    \mathcal{L}_{alignment} = MSE(S_{A}, S_{B}) ,
\end{equation*}

\paragraph{\textsc{The symmetrical architecture}}
To align all languages and maximize data usage, we adopt a symmetrical approach that unmasks the tokens of \langA{} with \sentB{}, and vice versa, simultaneously. 
We thus create four instances of the encoder (with shared parameters). %
For each language, we have two versions of each sentence: one heavily masked and one clean.
This allows us to generate two clean sentence vectors, \sentA{} and \sentB{}, which is essential for aligning representations between languages.
A non-symmetrical approach with only two encoders (one per language) would not produce the desired alignment as it would force the model to align a heavily masked sentence vector with a clean one, which is not ideal.

\paragraph{\textsc{The KoLeo loss}}
In preliminary experiments, we noticed that our representations exhibited more anisotropy than those learned with contrastive approaches. This has been shown to impact the quality of the representations~\citep{godeyAnisotropyInherentSelfAttention2024}.
Inspired by DINOv2~\citep{oquab2024dinov}, we employ the KoLeo loss \citep{sablayrolles2018spreading} to encourage sentence representations to spread out evenly in the latent space.
The KoLeo loss is based on the Kozachenko-Leonenko differential entropy estimator (see \citet{koleo}). 
We define below the KoLeo loss, $L_{KoLeo}$, for a set of $n$ representations, as well as the symmetrical version, $L_K$, we use to train our models:
$$
\mathcal{L}_{K} = \mathcal{L}_{KoLeo}(S_{A}) + \mathcal{L}_{KoLeo}(S_{B})  \ \ \ \mathrm{with} \ \ \ \mathcal{L}_{KoLeo} = - \frac{1}{n} \sum^{n}_{i=1} log(d_{n,i})
$$
where $d_{n,i} = min_{j \neq i} \parallel x_i - x_j \parallel$ is the distance between $x_i$ and its  nearest point in the batch.

Our training loss is a weighted combination of all previous losses:
\begin{equation*}
    \mathcal{L}_{MEXMA} = \alpha \cdot \mathcal{L}_{alignment} + \beta \cdot \mathcal{L}_{mlm} + \gamma \cdot \mathcal{L}_{K}
\end{equation*}
where $\alpha$, $\beta$ and $\gamma$ are hyper-parameters that control the weight of each loss term.
To show that MEXMA can be used on top of existing alignment approaches, we provide  in \Cref{subsecion:contrastive_loss} experimental results when replacing the MSE alignment loss in MEXMA with a contrastive loss.

\subsection{Experimental setup} \label{section:experimental_config}
\paragraph{\textsc{Encoder backbone}}
As our encoder, we utilize a modified version of the XLM-RoBERTa model \citep{conneau-etal-2020-unsupervised} provided by HuggingFace that uses a more efficient attention (details in \Cref{appendix:experimental_setup}).
Our sentence representation from the encoder is obtained via the CLS embedding of the last layer, without any further processing.
\paragraph{\textsc{Training data}}
Our training dataset is a subset of the NLLB-200 corpus \citep{nllbteam2022languageleftbehindscaling}, which comprises 200 languages. 
We cover 81 languages, utilizing only publicly available data, all sourced from Opus \citep{tiedemann-2012-parallel}.
The specific languages used are listed in \Cref{appendix:language_info}.
We always train using one sentence in English associated with its translation in one of the remaining 80 languages.
The dataset consists of a combination of human-translated and synthetic data, where we attempt to impose a minimum of 15 million sentences per language.
For languages with limited human-annotated data, we supplemented the dataset with mined data from NLLB \citep{schwenk2020ccmatrixminingbillionshighquality, fan2020englishcentricmultilingualmachinetranslation, nllbteam2022languageleftbehindscaling} to reach the 15 million sentence threshold.
Conversely, to ensure that our dataset is somewhat balanced across languages, for languages with abundant human-annotated data, we capped the dataset at 25 million sentences per language. 
The datasets used are detailed in \Cref{tab:appendix/datasets_used_to_train}.

We provide additional details about the parameters and configurations of our model in \Cref{appendix:experimental_setup}.

%% file: sections/evaluation.tex
\section{Results} \label{section:results}
To assess the quality and alignment of our embeddings, we evaluate them on a range of tasks. 
These tasks fall into two categories: mining tasks and other downstream tasks.
Mining tasks measure how aligned our representations are across languages, while downstream tasks evaluate the generalization power and overall quality of our embeddings.

\subsection{Multilingual alignment through mining}
We evaluate on three alignment tasks, namely xsim\footnote{\url{https://github.com/facebookresearch/LASER/tree/main/tasks/xsim}}, xsim++ \citep{chen-etal-2023-xsim} and BUCC \citep{zweigenbaum:hal-01898360, zweigenbaum-etal-2017-overview}.
xsim and BUCC are composed of sentences translated in many languages, and the goal is to be able to retrieve the correct translation of a query sentence.
xsim++ extends this task by introducing variations in the existing sentences in English, creating hard negatives that are difficult to distinguish from the correct sentence.
We follow \citet{heffernan-etal-2022-bitext} and do not evaluate on Tatoeba because of the small amount of data available for some language pairs and the low-quality translations created by non-professional volunteers.

xsim and xsim++ use a margin-based similarity approach \citep{artetxe-schwenk-2019-margin}.
We use the same setup as described in \citet{heffernan-etal-2022-bitext, duquenne2023sonar}.
For BUCC, the similarity is the cosine similarity as is commonly done.
The xsim and xsim++ scores are the error rate of wrongly aligned sentences in the test set.
For BUCC, the score is the F1 score of the alignment, computed using the MTEB benchmark \citep{muennighoff-etal-2023-mteb}.

BUCC evaluates on 4 languages: German, French, Russian and Chinese.
The detailed results (per language) are available in \Cref{appendix:mteb_datasets}.
We evaluate our model using xsim and xsim++ on the FLORES200 dataset, covering the 81 languages supported by our model (listed in \Cref{appendix:language_info}).
For fairer comparison, we also report results for the 72 languages supported by LaBSE, SONAR, and MEXMA ("o-xsim"), and separately for the 34 languages common to DAP and the other models ("d-xsim").
\begin{table}[]
    \centering
    \begin{tabular}{l|l|l|l|l|l|l|l}
        \hline
        \small{Model} & \small{xsim \scriptsize{$\downarrow$}} & \small{xsim++ \scriptsize{$\downarrow$}} & \small{BUCC \scriptsize{$\uparrow$}} & \small{o-xsim \scriptsize{$\downarrow$}} & \small{o-xsim++ \scriptsize{$\downarrow$}} & \small{d-xsim \scriptsize{$\downarrow$}} & \small{d-xsim++ \scriptsize{$\downarrow$}} \\
        \hline
        DAP   & -    & -     & 98.68 & -    & -     & 2.90 & 32.82\\
        SONAR & 0.09 & 12.08 & 98.25 & 0.08 & 11.68 & 0.04 & 10.55 \\
        LaBSE & 0.92 & 18.65 & 98.75 & 0.31 & 16.21 & 0.26 & 14.51 \\
        MEXMA & \textbf{0.06} & \textbf{9.60} & \textbf{98.93} & \textbf{0.05} & \textbf{9.01} & \textbf{0.02} & \textbf{8.26} \\
        \hline
    \end{tabular}
    \caption{Results in mining (\%). xsim and xsim++ are computed on 81 languages (FLORES200 dataset, X-eng pairs), with o-$\ldots$ columns showing results for 72 supported languages from LaBSE and d-$\ldots$ columns showing results for 34 languages supported by DAP. BUCC is computed with F1 on its 4 languages.}
    \label{tab:evaluation/mining_results}
\end{table}

The results are shown in \Cref{tab:evaluation/mining_results}.
MEXMA outperforms previous SOTA on all three benchmarks, showcasing the improved alignment achieved in our new approach.
The improvements in xsim and BUCC suggest that our approach improves the semantic alignment of the embeddings.
The large improvement in xsim++ (+2.48\% absolute improvement against the previous best model SONAR) also indicates the increased robustness of our model with regard to hard negatives, likely due to handling better lexical information.

\subsection{Downstream tasks}
To understand the quality of our embeddings and how generic they are, we evaluate them on several tasks from the MTEB benchmark \citep{muennighoff-etal-2023-mteb}.
We report the averaged results for each language.
For the full list of results for every task, see~\Cref{appendix:mteb_datasets}.

\paragraph{\textsc{Single sentence classification}}
We evaluate our model's classification performance on two benchmarks. 
First, the SentEval suite \citep{conneau2018senteval} is used to assess the performance across various tasks in English. We only evaluate on the tasks considered in LaBSE.
Second, we evaluate the multilingual classification capabilities using the available datasets from the MTEB benchmark.
\Cref{tab:evaluation/classification_results} shows the aggregated results. %
We can see that MEXMA outperforms all baseline models on average, and more specifically gains +2.33\% when compared with SONAR.
\begin{table}[]
    \centering
    \begin{tabular}{l|l|l|l|l|l|l|l|l}
        \hline
        Model & average & SentEval & en & zh & fr & da & nb & pol \\
        \hline
DAP   &    61.80 &     78.18 &     66.35 &   67.46 &    63.76 &     52.27 &     51.58 &     53.03 \\
SONAR &    63.02 &     85.82 &     65.63 &   63.13 &    61.88 &     54.01 &     55.59 &     55.09 \\
LaBSE &    62.77 &     85.63 &     66.75 &\bd{68.69}&    62.05 &     49.53 &     50.76 &     56.00 \\
MEXMA & \bd{65.35} & \bd{86.38} & \bd{68.20} &  66.25 & \bd{66.07} & \bd{55.38} & \bd{58.08} & \bd{57.09} \\
        \hline
    \end{tabular}
    \caption{Classification results, reported as accuracy (\%), on SentEval and MTEB (last 6 columns), averaged across languages. Full results in \Cref{appendix:mteb_datasets}.}
    \label{tab:evaluation/classification_results}
\end{table}

\paragraph{\textsc{Pairwise sentence Classification}}
We further evaluate on the pair classification task.
This task consists in classifying two sentences, e.g. determining if a pair of sentences are duplicates or not.
The metric, as reported in MTEB, is the Average Precision (AP) based on the distance between sentence representations.
The results are in \Cref{tab:evaluation/pair_classification_results}.
\begin{table}[]
    \centering
    \begin{tabular}{l|l|l|l|l}
        \hline
        Model & average & en & zh & fr \\
        \hline
        DAP   &           66.01  &         63.87  &       	61.12  &         73.03  \\
        SONAR &           69.70  &         70.73  &         60.80  &         77.57  \\
        LaBSE &           68.47  &         69.75  &         61.95  &         73.70  \\
        MEXMA &   \textbf{71.55} & \textbf{74.39} & \textbf{62.12} & \textbf{78.13} \\
        \hline
    \end{tabular}
    \caption{Pair classification results, reported as average precision (\%), on MTEB, averaged across languages. Full results in \Cref{appendix:mteb_datasets}.}
    \label{tab:evaluation/pair_classification_results}
\end{table}
MEXMA consistently outperforms all baselines on average, by at least +1.85\%.
These results, combined with our single sentence classification results, suggest that our model can effectively encode the relevant information in the sentence vectors.

\paragraph{\textsc{Semantic Textual Similarity (STS)}}
\begin{table}[]
    \centering
    \begin{tabular}{l|l|l|l|l|l}
        \hline
        Model & avg & eng & zh & fr & pl \\
        \hline
        DAP   & 59.39 & 67.45 & 45.31 & 67.74 & 57.06 \\
        SONAR & 58.04 & 67.24 & 42.15 & 65.60 & 57.17 \\
        LaBSE & \textbf{64.65} & \textbf{70.93} & 47.50 & \textbf{74.33} & \textbf{65.82} \\
        MEXMA & 63.99 & 70.62 & \textbf{51.56} & 70.10 & 63.67 \\
        \hline
    \end{tabular}
    \caption{STS results, reported as Spearman correlation (\%), on MTEB, averaged across languages. Full results in \Cref{appendix:mteb_datasets}.}
    \label{tab:evaluation/sts_results}
\end{table}
The STS task evaluates the model's ability to replicate human judgments on sentence similarity.
The metric, as reported in MTEB, is the Spearman correlation based on distance.
The results are in \Cref{tab:evaluation/sts_results}.
We can see that LaBSE outperforms all other methods, and in particular MEXMA by 0.66\%. 
MEXMA outperforms SONAR (+5.95\%) and DAP (+4.6\%).
The results indicate that the contrastive loss better suits the STS task, given that this is the only task where DAP is able to outperform SONAR, and where LaBSE outperforms MEXMA.

%% file: sections/experiments.tex
\section{Ablations and Analyses}
In this section, we conduct a comprehensive analysis of our MEXMA architecture, examining the impact of its individual components, how it scales with varying model and data sizes, and its potential to improve other alignment approaches.
We also examine the characteristics of the token embeddings and sentence representations learned by our model.
\subsection{Model components} \label{section:ablation_token_level_grads}
\begin{table}[]
    \centering
    \begin{tabular}{l|l|l|l}
        \hline
        component & xsim \scriptsize{$\downarrow$} & xsim++ \scriptsize{$\downarrow$} & SentEval \scriptsize{$\uparrow$} \\
        \hline
        Only sentence-level grads \circled{1} & 0.15 & 11.37 & 85.06 \\
        $+$ Token-level grads \circled{2} & 0.10 \tiny{\textcolor{green}{$\downarrow$0.05}} & 9.67 \tiny{\textcolor{green}{$\downarrow$1.7}} & 85.98 \tiny{\textcolor{green}{$\uparrow$0.92}} \\
        $+$ KoLeo loss \circled{3} - MEXMA & 0.06 \tiny{\textcolor{green}{$\downarrow$0.04}} & 9.60 \tiny{\textcolor{green}{$\downarrow$0.07}} & 86.38 \tiny{\textcolor{green}{$\uparrow$0.4}} \\
        \hline
    \end{tabular}
    \caption{\label{tab:experiments/short_model_ablation}Ablation study of the different components of the model. All experiments are conducted with the final parameters of the model, as reported in \Cref{section:experimental_config}.}
\end{table}

In \Cref{tab:experiments/short_model_ablation} we ablate the impact of having direct token-level gradients in MEXMA.
In model \circled{1}, we have all of MEXMA's components, as covered in \Cref{section:methodology}, without the KoLeo loss.
However, the gradients from the unmasking task are only back propagating through the sentence representations back to the encoder, and are deactivated for the individual tokens the encoder outputs, i.e. in the $\mathcal{L}_{mlm}$ mentioned in \Cref{section:methodology}, $\hat{A}$/$\hat{B}$ have no gradients flowing back to the encoder.
This model already achieves results that are competitive with current state of the art, but does not outperform them.
However, if we allow the gradients to flow through the tokens directly, model \circled{2}, we are able to outperform the current state-of-the-art.
As we hypothesized, adding updates on the tokens directly, coupled with the sentence updates largely improves results across all tasks.
Additionally, we also show that adding the KoLeo loss, model \circled{3}, also slightly improves results across all tasks.
The ablation on all components of the model is provided in \Cref{appendix:ablations}.

\subsection{Contrastive alignment loss} \label{subsecion:contrastive_loss}
\begin{table}[]
    \centering
    \begin{tabular}{l|l|l|l}
        \hline
        Model & xsim \scriptsize{$\downarrow$} & xsim++ \scriptsize{$\downarrow$} & SentEval \scriptsize{$\uparrow$} \\
        \hline
        Contrastive XLM-RoBERTa & 0.13 & 33.30 & 85.5 \\
        Contrastive MEXMA without MLM token-level gradients & 0.13  & 12.78 & 85.86 \\
        Contrastive MEXMA & 0.12 & 10.93 &  85.94 \\
        \hline
    \end{tabular}
    \caption{Using contrastive loss as the alignment loss in MEXMA.}
    \label{tab:experiments/contrastive_alignment}
\end{table}
To further assess the improvements given by the direct token updates in MEXMA, and understand MEXMA's scalability to other alignment approaches, we replaced our alignment loss, MSE, with a contrastive loss (also dropping the KoLeo loss).
We used a siamese network with XLM-RoBERTa-large trained on the symmetric cross-entropy loss (InfoNCE from \cite{oord2019representationlearningcontrastivepredictive}) as the baseline model, having an architecture similar to LaBSE \citep{feng-etal-2022-language}.
Our training used a batch size of 1.2k, with the rest of the parameters the same as reported in \Cref{section:experimental_config}.
The results are presented in \Cref{tab:experiments/contrastive_alignment}.
Our baseline model performed well on xsim and SentEval but struggled with xsim++.
Switching to the MEXMA architecture without token-level gradients, as done in model \circled{1} in \Cref{section:ablation_token_level_grads}, improved performance, already close to state-of-the-art xsim++ performance.
Moreover, incorporating token-level gradients, allowing the full MEXMA architecture with contrastive loss, as done in model \circled{2} in \Cref{section:ablation_token_level_grads}, resulted in competitive performance, already outperforming previous approaches in SentEval and xsim++.
This demonstrates the positive impact of direct token-level gradients and shows that MEXMA can be easily integrated with existing alignment approaches, such as contrastive learning, to improve their results.

\subsection{Model and data sizes}
\begin{table}[]
    \centering
    \begin{tabular}{l|l|l|l|l|l|l}
        \hline
         Model & \#parameters &	xsim \scriptsize{$\downarrow$} & xsim++ \scriptsize{$\downarrow$} &	SentEval \scriptsize{$\uparrow$} &  d-xsim \scriptsize{$\downarrow$} & d-xsim++ \scriptsize{$\downarrow$} \\
         \hline
            DAP        & 277M &      &       & 78.18 & 2.90 & 32.82 \\
            \textbf{MEXMA-base} & 277M & 0.13 & 13.03 & 85.30 & 0.06 & 11.01 \\
            LaBSE & 471M & 0.92 & 18.65 & 85.63 & 0.26 & 14.51 \\
            \textbf{MEXMA} & 559M & \textbf{0.06} & \textbf{9.60} & \textbf{86.38} & \textbf{0.02} & \textbf{8.26} \\
            SONAR & 766M & 0.09 & 12.08 & 85.82 & 0.04 & 10.55 \\
         \hline
    \end{tabular}
    \caption{Model size comparison. MEXMA-base is based on the XLM-RoBERTa-base, and MEXMA is based on XLM-RoBERTa-Large. The d-xxx columns are computed on 34 languages supported by DAP.}
    \label{tab:experiments/model_size}
\end{table}

\Cref{tab:experiments/model_size} shows how our model's results scale with the model size.
We train two models, MEXMA-base with 277M parameters, based on XLM-RoBERTa-base, and MEXMA with 559M parameters, based on XLM-RoBERTa-large.
It is possible to see that even the smaller model (277M parameters) outperforms LaBSE (471M parameters), on both xSIM and xSIM++, and gets a close result in SentEval, with a 0.3\% decrease in performance, with ~58.81\% of the size.
This smaller model also gets surprisingly close to the results of SONAR, which has 766M parameters, i.e. $\approx$2.77 times its size.
These results show that our approach works on smaller and larger models, and it seems to enable quite powerful small models, due to our stronger training signal.
Our larger model, MEXMA, with $\approx$73\% the size of SONAR, is able to largely outperform it across all tasks.

\begin{table}[]
    \centering
    \begin{tabular}{l|l|l|l||l|l|l}
        \hline
         \small{Model} &	\small{81 xsim \scriptsize{$\downarrow$}} & \small{81 xsim++ \scriptsize{$\downarrow$}} & \small{SentEval \scriptsize{$\uparrow$}} & \small{90 xsim \scriptsize{$\downarrow$}} & \small{90 xsim++ \scriptsize{$\downarrow$}} & \small{SentEval \scriptsize{$\uparrow$}} \\
         \hline
         SONAR & 0.09 & 12.08 & 85.82 & \textbf{0.05} & 11.42 & 85.82 \\
         MEXMA & \textbf{0.06} & \textbf{9.60} & \textbf{86.38} & \textbf{0.05} & \textbf{9.06} & \textbf{86.64} \\
         \hline
    \end{tabular}
    \caption{
        Training data size comparison. We train MEXMA on either 81 languages, or 90 languages. 
        See \Cref{appendix:language_info} for the list of covered languages. 
    }
    \label{tab:experiments/training_data}
\end{table}

To investigate the impact of training data, we conducted experiments using two different language subsets of the FLORES200.
We trained separate MEXMA models on each subset, using the same hyperparameters as reported in \Cref{section:experimental_config}. 
For comparison, we evaluated the publicly available SONAR model, which was trained on all available 200 languages, on both
language
subsets.
The results, presented in \Cref{tab:experiments/training_data}, demonstrate that MEXMA outperforms SONAR on both subsets, highlighting the adaptability and robustness of our approach to varying training data. 

\subsection{Masking ratio}
NLP models typically use masking percentages around 15\%, whereas vision papers have explored much higher masking ratios, ranging from 40\% in BEiT \citep{bao2022beit} to as high as 90\% in MAE \citep{He_2022_CVPR} and V-JEPA \citep{bardes2024revisiting}, usually aligning \textit{augmentations}.
For text, there is less redundancy and the representations are more information-dense.
In our case, we are aligning the same sentence in several languages, which can be viewed as \textit{augmentations} of a pivot sentence, i.e. the sentence in English.
We need to know how much we can mask, to make the unmasking task hard, but to not deteriorate the performance of our encoder.
\Cref{tab:experiments/masking_ratio} shows the results we obtained for the different masking ratios.
The range 30\%-60\% seems to be the best operating region.
We selected 40\% for all experiments conducted in this paper, since it had the best balance between alignment and classification.
More information is provided in \Cref{appendix:ablations}.

\subsection{Token embeddings analysis}
\begin{table}[]
    \centering
    \begin{tabular}{l|l|l|l|l}
        \hline
        Model & \% other & \% same language & \% same sentence & \% translation \\
        \hline
        XLM-RoBERTa	       & 1.19 & 63.89 & 2.65  & 32.28 \\
        LaBSE	           & 0.00 & 0.13  & 42.33 & 57.54 \\
        DAP                & 0.00 &	0.66  & 20.11 &	79.23 \\
        No-tok-MEXMA       & 0.13 & 0.40  & 11.90 & 87.57 \\
        NLLB	           & 0.40 & 3.17  & 1.72  & 94.71 \\
        SONAR	           & 0.00 & 0.13  & 0.20  & 99.67 \\
        MEXMA	           & 0.26 & 1.33  & 0.53  & 97.88 \\
        \hline
    \end{tabular}
    \caption{Result of the token matching analysis.}
    \label{tab:analysis/token_matching_breakdown}
\end{table}
Sentence vectors are pooled representations of their tokens.
In this section, we investigate the information encoded in the tokens from the last layer across different models.
Our goal is to determine whether the tokens primarily convey semantic, lexical, and/or contextual information.
Although these categories can be intertwined, understanding the dominant characteristics of each model's tokens provides valuable insights into their behavior.

To gain insight into the information encoded in individual tokens, we examined their nearest neighbors in the contextual embedding space.
We categorized these neighboring tokens into four groups based on the sentence they belong to.
\emph{Same language}: the matched token is the same token in a different sentence in the same language, which means that it encodes lexical information.
\emph{Same sentence}: the token matches another one in the same sentence, meaning the tokens representations are heavily influenced by the context.
\emph{Translation}: the token matches its equivalent in a translation of the original sentence. It means that the tokens are aligned across languages.
\emph{Other}: tokens that do not belong to previous classes.

We conducted these experiments by encoding all tokens from all sentences of the 81 languages (see \Cref{appendix:language_info} for the list) on the FLORES200 test set using each model. 
We randomly select three tokens among each of the first 250 English sentences of the dataset as query, and for each query, we retrieve the five closest tokens among all tokens of all sentences (but itself).
We analyze the properties of the sentence encoders 
as well as some respective backbones, XLM-RoBERTa (used to initialize MEXMA) and NLLB-200 (used for SONAR).
For the sake of comparison, we also examine "no-tok-MEXMA", a variant of MEXMA that does not use token-level gradients during training.
The statistics are shown in \Cref{tab:analysis/token_matching_breakdown}.

Our analysis reveals distinct characteristics for the considered models and we can cluster them in three different overall behaviours.
XLM-RoBERTa exhibits strong lexical relationships (high \emph{same language} percentage) but weaker semantic and contextual relations. %

LaBSE, DAP and no-tok-MEXMA show higher semantic capabilities as shown by the larger \emph{translation} rate. %
However, we can also observe a high percentage of matches with adjacent tokens (\emph{same sentence} column), indicating that those models encode a very large amount of context in their tokens.

NLLB, SONAR and MEXMA have strong cross-lingual semantic capability as shown by the very high percentage in the \emph{translation} column. 
This is expected as SONAR and NLLB were trained to perform translation, and MEXMA cross-lingual unmasking. %
Notice that for SONAR and MEXMA, this cross-lingual token level alignment is guided by the decoding using the sentence representation as context (and additionally the direct token-level gradients for MEXMA).

Note also that LaBSE and DAP are the only models trained with a sentence-level contrastive loss, and even though DAP has an additional loss to enforce the semantic alignment of the tokens, it does not manage to achieve the same alignment as SONAR and MEXMA.

Notably, comparing the backbones NLLB and XLM-RoBERTa, we can see that the former exhibits more semantical tokens than the latter, as shown by its higher \textit{translation} rate and lower \textit{same sentence} rate, which can be attributed to its translation-based pre-training that enhances semantic properties and cross-lingual alignment.
SONAR, which starts from NLLB, also matches translated tokens with a high rate, \textgreater{}99\%, 
but does not encode a lot of lexical information (low \textit{same language} rate). %
MEXMA also matches translated tokens very frequently, but additionally displays more lexicality (higher \textit{same language} rate) and increased semantic robustness (higher \textit{other} rate).
We verified MEXMA's \textit{other} matches, and the matched tokens belong to sentences in other languages that are not translations of the original one, matching the translated token, exhibiting semantic properties.
All properties displayed by MEXMA allow it to create sentence representations that inherit those same properties, allowing it to outperform other models on downstream tasks.
We provide examples to illustrate the behavior of the models in \Cref{section:appendix/token_level_analysis}.

\subsection{Sentence vector analysis} \label{section:sentence_vector_analysis}
The sentence representations are created by combining the tokens' representations
in various ways (average or CLS/attention pooling).
Previous section looked at the properties encoded in those tokens, and in this section, we aim to look at how those representations are combined to create the sentence embedding.

For SONAR, the attention weight distribution is uniform, given that SONAR averages the tokens to create their sentence representation.
MEXMA and LaBSE both use a CLS token to perform pooling over the tokens in the sentence.

MEXMA's and LaBSE's attention distribution are rather different, with LaBSE having a more uniform attention distribution across its tokens, and MEXMA having a more skewed representation.
We verify this by computing the average entropy of the attention probabilities in the last layer given by the CLS token, for both models on the test set of the FLORES200, in the languages supported by both LaBSE and MEXMA.
LaBSE gets an entropy of $\approx$ 3.4, while MEXMA gets an entropy of $\approx$ 2.5.
The entropy values obtained for LaBSE and MEXMA are difficult to interpret in absolute terms, but the relative difference between them is informative.
Specifically, LaBSE exhibits a higher entropy compared to MEXMA, suggesting that it has a more uniform distribution of attention probabilities.
We provide examples of the distributions in \Cref{appendix:attn_distribution_tokens}.

\begin{table}[]
    \centering
    \begin{tabular}{l|l|l|l|l}
        \hline
        Model & xsim \scriptsize{$\downarrow$} & xsim++ \scriptsize{$\downarrow$} & STS \scriptsize{$\uparrow$} & Classification \scriptsize{$\uparrow$} \\
        \hline
        Uni LaBSE & 2.02 & 20.73 & 63.50 & 58.03 \\
        Uni MEXMA & 0.19 & 18.21 & 54.24 & 56.98 \\
        \hline
        CLS LaBSE & 0.92 & 18.65 & 64.65 & 62.77 \\
        CLS MEXMA & 0.06 & 9.60 &  63.99 & 65.35 \\
        \hline
        \hline
        $\Delta$ LaBSE & -119.65 & -11.19 & +1.78 &  +7.55 \\ 
        $\Delta$ MEXMA & -212.50 & -89.73 & +15.24 & +12.81 \\ 
        \hline
    \end{tabular}
    \caption{Downstream results for LaBSE and MEXMA, using both a uniform attention distribution (Uni xxx in the table), and the CLS distribution (CLS xxx in the table). The last two rows provide the delta between the uniform and CLS distributions, in relative terms. Classification and STS results are across all datasets mentioned under \Cref{appendix:mteb_datasets}.}
    \label{tab:analysis/avg_of_tokens_as_sent_rep_analysis}
\end{table}
We perform an additional analysis, where we push the uniformity of the sentence representation to the extreme, by using the average of tokens as our sentence representation.
By doing this for both MEXMA and LaBSE, 
we aim to understand the importance/impact of the distribution for each model.
The results are provided in \Cref{tab:analysis/avg_of_tokens_as_sent_rep_analysis}.
The deltas are computed in terms of relative change from the uniform to the CLS representation.
We can see that for all tasks, MEXMA has a larger change in performance compared to LaBSE, showing that indeed since our representations are more skewed, we suffer more from an increase in uniformity of the distribution.
For those tasks, it is noticeable that MEXMA having a uniform distribution, will loose its ability to focus on the important tokens, decreasing its results.
For LaBSE the decrease is not as accentuated, since it was already not focusing as much on the important tokens with its more uniform CLS pooling.

%% file: sections/conclusion.tex
\section{Conclusion}
We introduced MEXMA, a novel multilingual alignment technique that leverages both token-level and sentence-level objectives.
We show that integrating token-level objectives into the training of cross-lingual sentence encoders greatly improves their sentence representation quality, achieving new state-of-the-art results in bitext mining and other downstream tasks.
We additionally validate these improvements via ablations.
Notably, MEXMA also achieves strong token alignment across languages and effectively encodes meaningful information within each token.
Since the sentence representation is built from these tokens, as we analysed, this leads to better sentence representations.
Looking ahead, we plan to explore MEXMA's scalability to more languages, and potentially modalities.

%% file: sections/acknowledgments.tex
\subsubsection*{Acknowledgments}
We would like to thank Belen Alastruey for the help with the figures, and several technical discussions.
We would also like to thank Timothee Darcet and Robin San Roman for several technical discussions.

%% file: sections/appendix_experimental_setup.tex
\section{Experimental Setup} \label{appendix:experimental_setup}
\subsection{Encoder backbone}
The available implementation of XLM-RoBERTa in HuggingFace employs an inefficient attention mechanism, which we have modified to incorporate the memory-efficient attention from xFormers \citep{xFormers2022}. 
This modification was necessary due to the random batching process used in our training, which results in a significant amount of padding and increased computational cost. 
To address this issue and eliminate padding, we have employed the BlockDiagonalMask \footnote{\url{https://facebookresearch.github.io/xformers/components/ops.html\#xformers.ops.fmha.attn_bias.BlockDiagonalMask}}, which through custom CUDA kernels, avoids computations in padding altogether.
With this change we are able to increase our batch size in each GPU by a factor of $\approx$ 8.

\subsection{Unmasking head}
For the unmasking head, we use 6 transformer layers, also leveraging the memory-efficient attention.

\subsection{Compute and training length}
Our models were trained on a single node of 8 A100 GPUs.
Each GPU had a batch size of 150, totalling 1,200 batch size across all GPUs.
We accumulated two gradients, making our effective batch size 2,400.
We trained our models for 300k steps.

\subsection{Losses}
Our models were trained with $\alpha=1$, $\beta=\frac{1}{2}$ and $\gamma=\frac{0.01}{2}$.

\subsection{Training parameters}
We utilize the AdamW optimizer for our training process.
The learning rate is linearly increased from 1e-5 for the 300k steps.
To optimize memory usage, we employ mixed precision training, where the model is stored in float32, while most computations are performed in float16. 
The maximum sequence length for our input data is set to 200 tokens. 
Finally, we apply a masking ratio of 40\% to the input data.

%% file: sections/appendix_ablations.tex
\section{Ablations} \label{appendix:ablations}

\subsection{Model components}
\begin{table}[h]
    \centering
    \begin{tabular}{l|l|l|l}
        \hline
        component & xsim \scriptsize{$\downarrow$} & xsim++ \scriptsize{$\downarrow$} & SentEval \scriptsize{$\uparrow$} \\
        \hline
        Non-symmetrical \circled{1} & 0.09 & 14.75 & 84.68 \\
        $+$ Symmetrical architecture \circled{2} &	0.09 \tiny{\textcolor{blue}{0.00}} & 14.39 \tiny{\textcolor{green}{$\downarrow$0.36}} & 84.83 \tiny{\textcolor{green}{$\uparrow$0.15}}  \\
        $+$ Alignment loss (clean to dirty alignment) \circled{3} & 0.21 \tiny{\textcolor{red}{$\uparrow$0.12}} & 12.09 \tiny{\textcolor{green}{$\downarrow$2.3}} & 85.61 \tiny{\textcolor{green}{$\uparrow$0.78}}\\
        $+$ Clean to clean alignment \circled{4} & 0.15 \tiny{\textcolor{green}{$\downarrow$0.06}} & 11.37 \tiny{\textcolor{green}{$\downarrow$0.72}} & 85.06 \tiny{\textcolor{red}{$\downarrow$0.55}}\\
        $+$ Token-level grads \circled{5} & 0.10 \tiny{\textcolor{green}{$\downarrow$0.05}} & 9.67 \tiny{\textcolor{green}{$\downarrow$1.7}} & 85.98 \tiny{\textcolor{green}{$\uparrow$0.92}} \\
        $+$ KoLeo loss \circled{6} - MEXMA & 0.06 \tiny{\textcolor{green}{$\downarrow$0.04}} & 9.60 \tiny{\textcolor{green}{$\downarrow$0.07}} & 86.38 \tiny{\textcolor{green}{$\uparrow$0.4}} \\
        \hline
    \end{tabular}
    \caption{\label{tab:ablations/full_model_ablation}Ablation study of the different components of the model. All experiments are conducted with the final parameters of the model, as reported in \Cref{section:experimental_config}.}
\end{table}

In \Cref{tab:ablations/full_model_ablation}, we ablate the different components of our architecture described in \Cref{section:methodology}.
We briefly explain each entry in the table.
Model \circled{1} has only two encoder instances, one for each language, where one of the inputs is masked, and the other is left clean.
The token unmasking is performed with the clean sentence representation as context.
The languages are randomly swapped at every new sample, to eliminate potential biases.
The gradients from the unmasking task are only propagated back to the encoder via the sentence representation, and there is no gradient propagation from the individual tokens back to the encoder.
There is also neither alignment nor koleo losses.
Model \circled{2} adds two additional encoder instances, totalling four instances, two for each language, where now each language has its clean and masked input.
This allows to unmask \langA{} with \langB{}, and vice-versa, and will also allow (once added) to align two clean sentence representations.
Model \circled{3} adds the alignment loss, but it is performed between the masked sentence representation of \langA{} and the clean sentence representation of \langB{}, to better emphasize the advantages of having a symmetrical architecture with an alignment loss between two clean representations.
Model \circled{4} then changes the alignment loss to be performed between the two clean sentence representations of each language.
In model \circled{5} we allow the gradients from the unmasking to be propagated to the encoder via each individual token, as well as its sentence representation.
Finally, model \circled{6} adds the KoLeo loss.

The results indicate that each component always enhances performance on at least two out of the three tasks.
Notably, the alignment loss, \circled{3}-\circled{4}, and token-level gradients, \circled{5}, emerge as the most critical components.
More precisely, the alignment loss yields substantial improvements on two tasks while also resulting in a notable decline in performance on another task.
In contrast, the token-level gradients consistently provide significant performance gains across all three tasks.

\subsection{Masking ratio}
\begin{table}[h]
    \centering
    \begin{tabular}{l|l|l|l}
        \hline
         Masking \% & xSIM \scriptsize{$\downarrow$} & xSIM++ \scriptsize{$\downarrow$} & SentEval \scriptsize{$\uparrow$} \\
         \hline
        20\% & 0.06 & 10.50 & 85.87 \\
        30\% & 0.06 & 9.82 & 86.00 \\
        40\% & 0.06 & 9.60 & 86.38 \\
        50\% & 0.07 & 9.56 & 86.37 \\
        60\% & 0.08 & 9.79 & 86.13 \\
        70\% & 0.09 & 10.65 & 86.41 \\
        80\% & 0.10 & 12.81 & 85.85 \\
        90\% & 0.11 & 14.62 & 84.99 \\
        \hline
    \end{tabular}
    \caption{The model performance across different masking ratios.}
    \label{tab:experiments/masking_ratio}
\end{table}
Classical NLP masked encoders like BERT use a small masking percentage, usually $\approx$ 15\%, without aligning any \textit{augmentations}.
Recent vision approaches use much higher masking percentages.
BEiT \citep{bao2022beit} was one of the first masked image modelling (MIM) approaches, in a BERT-style training, and masked 40\%.
MAE \citep{He_2022_CVPR} is another BERT-like model for images, and masks 75\%, but shows that even masking 80\% or 90\% still achieves good results.
DINO v2 \citep{oquab2024dinov} and I-BOT \citep{zhou2021ibot} mask between 10\%-50\% in a block-wise masking scenario, aligning \textit{augmentations}.
I-BOT can use  65\%-75\% masking ratio, if randomly masking (instead of block-wise masking).
For videos, V-JEPA \citep{bardes2024revisiting} masks with a very high percentage of 90\%.
Recent textual approaches, namely RetroMAE experiment with masking percentages of up to 50$\sim$70\%, but this task will not update the actual encoder.

For MEXMA, since these masking gradients are updating our encoder, we need to strive for a balance where unmasking is hard, and cannot be done by the encoder and head, but also not too much that will deteriorate the representations of the encoder.
\Cref{tab:experiments/masking_ratio} shows the results we obtained for the different masking ratios.

%% file: sections/appendix_language_info.tex
\section{Language information appendix} \label{appendix:language_info}
In this section, we cover the languages used by our model.
The list of languages used to train our model is reported in \Cref{tab:appendix/80_languages_set}.
The list used to conduct the experiments with 90 languages is available in \Cref{tab:appendix/90_languages_set}.
\begin{table}[h]
    \centering
    \scalebox{0.9}{%
        \begin{tabular}{l|l||l|l}
            \hline
            FLORES200 code & Language & FLORES200 code & Language  \\
            \hline
            acm\_Arab & Mesopotamian Arabic & kan\_Knda & Kannada \\
            aeb\_Arab & Tunisian Arabic & kat\_Geor & Georgian \\
            afr\_Latn & Afrikaans & kaz\_Cyrl & Kazakh \\
            amh\_Ethi & Amharic & khm\_Khmr & Khmer \\
            ary\_Arab & Moroccan Arabic & kir\_Cyrl & Kyrgyz \\
            arz\_Arab & Egyptian Arabic & kor\_Hang & Korean \\
            asm\_Beng & Assamese & lao\_Laoo & Lao \\
            azb\_Arab & South Azerbaijani & mal\_Mlym & Malayalam \\
            azj\_Latn & Azerbaijani & mar\_Deva & Marathi \\
            bel\_Cyrl & Belarusian & mkd\_Cyrl & Macedonian \\
            ben\_Beng & Bengali & mya\_Mymr & Burmese \\
            bos\_Latn & Bosnian & nld\_Latn & Dutch \\
            bul\_Cyrl & Bulgarian & nno\_Latn & Norwegian \\
            cat\_Latn & Catalan & nob\_Latn & Norwegian Bokmål \\
            ces\_Latn & Czech & npi\_Deva & Nepali \\
            ckb\_Arab & Central Kurdish & pol\_Latn & Polish \\
            cym\_Latn & Welsh & por\_Latn & Portuguese \\
            dan\_Latn & Danish & ron\_Latn & Romanian \\
            deu\_Latn & German & rus\_Cyrl & Russian \\
            ell\_Grek & Greek & san\_Deva & Sanskrit \\
            eng\_Latn & English & sin\_Sinh & Sinhala \\
            epo\_Latn & Esperanto & slk\_Latn & Slovak \\
            est\_Latn & Estonian & slv\_Latn & Slovenian \\
            eus\_Latn & Basque & snd\_Arab & Sindhi \\
            fin\_Latn & Finnish & som\_Latn & Somali \\
            fra\_Latn & French & spa\_Latn & Spanish \\
            gla\_Latn & Scottish Gaelic & srp\_Cyrl & Serbian \\
            gle\_Latn & Irish & sun\_Latn & Sundanese \\
            glg\_Latn & Galician & swe\_Latn & Swedish \\
            guj\_Gujr & Gujarati & swh\_Latn & Swahili \\
            hau\_Latn & Hausa & tam\_Taml & Tamil \\
            heb\_Hebr & Hebrew & tel\_Telu & Telugu \\
            hin\_Deva & Hindi & tha\_Thai & Thai \\
            hrv\_Latn & Croatian & tur\_Latn & Turkish \\
            hun\_Latn & Hungarian & uig\_Arab & Uyghur \\
            hye\_Armn & Armenian & ukr\_Cyrl & Ukrainian \\
            ind\_Latn & Indonesian & urd\_Arab & Urdu \\
            isl\_Latn & Icelandic & vie\_Latn & Vietnamese \\
            ita\_Latn & Italian & xho\_Latn & Xhosa \\
            jav\_Latn & Javanese & zho\_Hant & Chinese (Traditional) \\
            jpn\_Jpan & Japanese &  \\ 
            \hline
        \end{tabular}%
    }
    \caption{81 languages set.}
    \label{tab:appendix/80_languages_set}
\end{table}

\begin{table}[]
    \centering
    \begin{tabular}{l|l||l|l}
        \hline
        FLORES200 code & Language & FLORES200 code & Language  \\
        \hline
        afr\_Latn & Afrikaans & kmr\_Latn & Kurdish (Kurmanji) \\
        als\_Latn & Albanian & kor\_Hang & Korean \\
        amh\_Ethi & Amharic & lao\_Laoo & Lao \\
        arb\_Arab & Arabic & lit\_Latn & Lithuanian \\
        asm\_Beng & Assamese & lvs\_Latn & Latvian \\
        azj\_Latn & Azerbaijani & mal\_Mlym & Malayalam \\
        bel\_Cyrl & Belarusian & mar\_Deva & Marathi \\
        ben\_Beng & Bengali & mkd\_Cyrl & Macedonian \\
        bos\_Latn & Bosnian & mya\_Mymr & Burmese \\
        bul\_Cyrl & Bulgarian & nld\_Latn & Dutch \\
        cat\_Latn & Catalan & nno\_Latn & Norwegian \\
        ces\_Latn & Czech & npi\_Deva & Nepali \\
        cym\_Latn & Welsh & ory\_Orya & Oriya \\
        dan\_Latn & Danish & pan\_Guru & Punjabi \\
        deu\_Latn & German & pbt\_Arab & Pashto \\
        ell\_Grek & Greek & plt\_Latn & Malagasy \\
        eng\_Latn & English & pol\_Latn & Polish \\
        epo\_Latn & Esperanto & por\_Latn & Portuguese \\
        est\_Latn & Estonian & prs\_Arab & Persian \\
        eus\_Latn & Basque & ron\_Latn & Romanian \\
        fin\_Latn & Finnish & rus\_Cyrl & Russian \\
        fra\_Latn & French & san\_Deva & Sanskrit \\
        gaz\_Latn & Oromo & sin\_Sinh & Sinhala \\
        gla\_Latn & Gaelic & slk\_Latn & Slovak \\
        gle\_Latn & Irish & slv\_Latn & Slovenian \\
        glg\_Latn & Galician & snd\_Arab & Sindhi \\
        guj\_Gujr & Gujarati & som\_Latn & Somali \\
        hau\_Latn & Hausa & spa\_Latn & Spanish \\
        heb\_Hebr & Hebrew & srp\_Cyrl & Serbian \\
        hin\_Deva & Hindi & sun\_Latn & Sundanese \\
        hrv\_Latn & Croatian & swe\_Latn & Swedish \\
        hun\_Latn & Hungarian & swh\_Latn & Swahili \\
        hye\_Armn & Armenian & tam\_Taml & Tamil \\
        ind\_Latn & Indonesian & tel\_Telu & Telugu \\
        isl\_Latn & Icelandic & tha\_Thai & Thai \\
        ita\_Latn & Italian & tur\_Latn & Turkish \\
        jav\_Latn & Javanese & uig\_Arab & Uyghur \\
        jpn\_Jpan & Japanese & ukr\_Cyrl & Ukrainian \\
        kan\_Knda & Kannada & urd\_Arab & Urdu \\
        kat\_Geor & Georgian & uzn\_Latn & Uzbek \\
        kaz\_Cyrl & Kazakh & vie\_Latn & Vietnamese \\
        khk\_Cyrl & Mongolian & xho\_Latn & Xhosa \\
        khm\_Khmr & Khmer & ydd\_Hebr & Yiddish \\
        kir\_Cyrl & Kyrgyz & zho\_Hans & Chinese (Simplified) \\
        zsm\_Latn & Malay & zho\_Hant & Chinese (Traditional) \\
        \hline
    \end{tabular}
    \caption{90 languages set}
    \label{tab:appendix/90_languages_set}
\end{table}

%% file: sections/appendix_datasets.tex
\section{Datasets}
In this section we report the data used to train our models.
\Cref{tab:appendix/datasets_used_to_train} reports all the datasets used to train the models.

\begin{table}[h]
    \centering
    \begin{tabular}{l|l|l}
        \hline
        Dataset & Source & Origin \\
        \hline
        bible-uedin & Opus    &    \citet{christodouloupoulos2015massively, tiedemann-2012-parallel} \\
        DGT & Opus    &            \citet{steinberger-etal-2012-dgt, tiedemann-2012-parallel} \\
        ECB & Opus    &            \citet{tiedemann-2012-parallel} \\
        EMEA & Opus    &           \citet{tiedemann-2012-parallel} \\
        EUbookshop & Opus    &     \citet{tiedemann-2012-parallel} \\
        infopankki & Opus    &     \citet{tiedemann-2012-parallel} \\
        memat & Opus    &          \citet{tiedemann-2012-parallel} \\
        OpenSubtitles & Opus    &  \citet{lison-tiedemann-2016-opensubtitles2016, tiedemann-2012-parallel}, Link: \url{opensubtitles.org} \\
        QED & Opus    &            \cite{abdelali-etal-2014-amara, tiedemann-2012-parallel} \\
        Tanzil & Opus    &         \citet{tiedemann-2012-parallel}, Link: \url{tanzil.net/trans} \\
        Tatoeba & Opus    &        \citet{tiedemann-2012-parallel} \\
        Ted20 & Opus    &          \citet{reimers-2020-multilingual-sentence-bert, tiedemann-2012-parallel} \\
        Tico19 & Opus    &         \citet{anastasopoulos-etal-2020-tico, tiedemann-2012-parallel} \\
        UNPC & Opus    &             \citet{ziemski-etal-2016-united, tiedemann-2012-parallel} \\
        Wikimedia & Opus    &      \citet{tiedemann-2012-parallel} \\
        NLLB mined & Opus    &     \citet{schwenk2020ccmatrixminingbillionshighquality, fan2020englishcentricmultilingualmachinetranslation, tiedemann-2012-parallel} \\
        \hline
    \end{tabular}
    \caption{Datasets used to train our models.}
    \label{tab:appendix/datasets_used_to_train}
\end{table}

%% file: sections/appendix_mteb_datasets.tex
\section{MTEB datasets} \label{appendix:mteb_datasets}
In this section, we report the scores for each task of the MTEB benchmark reported in \Cref{section:results}.
We report the scores per task, with every dataset used per task, and per language.
MEXMA is able to outperform the previous SOTA results on mining, while also improving the downstream results on classification and pair classification.
LaBSE outperforms all other models on STS. %
\subsection{Bitext Mining}
Results for mining are in \Cref{tab:appendix/mteb_bucc_results}, for the BUCC dataset.
We report the scores on the four available languages: German, French, Russian and Chinese.
\begin{table}[h]
    \centering
    \begin{tabular}{l|l|l|l|l}
        \hline
        LP    & DAP & SONAR & LaBSE & MEXMA \\
        \hline
        de-en & 99.45 & 98.82 & 99.35 & 99.52 \\
        fr-en & 98.58 & 98.09 & 98.72 & 98.98 \\
        ru-en & 97.74 & 97.37 & 97.78 & 98.06 \\
        zh-en & 98.96 & 98.72 & 99.16 & 99.18 \\
        \hline
    \end{tabular}
    \caption{BUCC results for each language pair (LP).}
    \label{tab:appendix/mteb_bucc_results}
\end{table}

\subsection{Classification}
Classification results for English are available in \Cref{tab:appendix/senteval_results}, for SentEval, and in \Cref{tab:appendix/mteb_eng_classification} for the English MTEB classification datasets.
Classification results for Chinese, French, Danish, Norwegian and Polish are reported in \Cref{tab:appendix/mteb_zh_classification}, \Cref{tab:appendix/mteb_fr_classification}, \Cref{tab:appendix/mteb_da_classification}, \Cref{tab:appendix/mteb_ng_classification}, \Cref{tab:appendix/mteb_pl_classification}, respectively.
MEXMA outperforms all other models on average.
\begin{table}[h]
    \centering
    \begin{tabular}{l|l|l|l|l}
        \hline
        Task & DAP & SONAR & LaBSE & MEXMA \\
        \hline
        Average & 78.18 & 85.82 & 85.63 & 86.38 \\
        MR &      74.33 & 81.23 & 78.89 & 80.14 \\
        SST2 &    81.88 & 86.49 & 83.64 & 86.16 \\
        TREC &    75.00 & 95.00 & 92.80 & 94.80 \\
        CR &      78.70 & 85.67 & 86.44 & 84.43 \\
        SUBJ &    91.83 & 93.70 & 93.14 & 94.27 \\
        MPQA &    78.86 & 89.38 & 89.66 & 89.41 \\
        MRPC &    66.67 & 69.28 & 74.84 & 75.42 \\
        \hline
    \end{tabular}
    \caption{SentEval results.}
    \label{tab:appendix/senteval_results}
\end{table}

\begin{table}[h]
    \centering
    \begin{tabular}{l|l|l|l|l}
        \hline
        Dataset & DAP & SONAR & LaBSE & MEXMA \\
        \hline
        Average & 66.35 & 65.63 & 66.75 & 68.20 \\
        AmazonCounterfactualClassification & 77.16 &  81.49 &  75.93 & 78.06 \\
        AmazonPolarityClassification & 65.73 &  62.73 &  68.95 & 64.96 \\
        AmazonReviewsClassification & 34.03 &  31.55 &  35.80 & 32.77 \\
        Banking77Classification & 71.83 &  73.50 &  69.85 & 75.14 \\
        ImdbClassification & 62.06 &  55.75 &  62.04 & 62.08 \\
        MTOPDomainClassification & 85.54 &  89.92 &  86.06 & 89.85 \\
        MTOPIntentClassification & 64.17 &  70.85 &  63.03 & 75.18 \\
        MasakhaNEWSClassification & 77.95 &  55.42 &  77.77 & 72.28 \\
        MassiveIntentClassification & 63.48 &  64.37 &  61.46 & 66.64 \\
        MassiveScenarioClassification & 68.75 &  69.05 &  66.41 & 70.38 \\
        ToxicConversationsClassification & 59.14 &  67.28 &  66.90 & 62.85 \\
        \hline
    \end{tabular}
    \caption{MTEB English classification results.}
    \label{tab:appendix/mteb_eng_classification}
\end{table}

\begin{table}[h]
    \centering
    \begin{tabular}{l|l|l|l|l}
        \hline
        Dataset & DAP & SONAR & LaBSE & MEXMA \\
        \hline
        Average & 67.46	& 63.13 & 68.69 & 66.25 \\
        AmazonReviewsClassification (zh) & 34.35 & 31.91 & 32.98 & 33.40 \\
        MassiveIntentClassification (zh-CN) & 71.99 & 62.08 & 63.86 & 74.41 \\
        MassiveScenarioClassification (zh-CN) & 65.45 & 68.88 & 70.85 & 65.28 \\
        JDReview & 71.54 & 69.59 & 79.13 & 70.73 \\
        MultilingualSentiment & 62.03 & 57.69 & 65.52 & 60.34 \\
        OnlineShopping & 85.03 & 75.64 & 85.62 & 80.09 \\
        Waimai & 81.82 & 76.12 & 82.85 & 79.52 \\
        \hline
    \end{tabular}
    \caption{MTEB Chinese classification results.}
    \label{tab:appendix/mteb_zh_classification}
\end{table}

\begin{table}[h]
    \centering
    \begin{tabular}{l|l|l|l|l}
        \hline
        Dataset & DAP & SONAR & LaBSE & MEXMA \\
        \hline
        Average & 63.76 & 61.88 & 62.05 & 66.07 \\
        AmazonReviewsClassification & 35.60 & 34.91 & 38.52 & 35.62 \\
        MTOPDomainClassification & 84.43 & 86.19 & 84.14 & 86.70 \\
        MTOPIntentClassification & 65.78 & 66.75 & 62.01 & 74.12 \\
        MassiveIntentClassification & 64.51 & 58.55 & 60.47 & 65.59 \\
        MassiveScenarioClassification & 68.50 & 63.02 & 65.1 & 68.31 \\
        \hline
    \end{tabular}
    \caption{MTEB French classification results.}
    \label{tab:appendix/mteb_fr_classification}
\end{table}

\begin{table}[h]
    \centering
    \begin{tabular}{l|l|l|l|l}
        \hline
        Dataset & DAP & SONAR & LaBSE & MEXMA \\
        \hline
        Average & 52.27 & 54.01 & 49.53 & 55.38 \\
        DanishPoliticalCommentsClassification & 36.44 & 37.59 & 38.69 & 38.75 \\
        LccSentimentClassification & 58.27 & 54.27 & 50.07 & 52.40 \\
        MassiveIntentClassification (da) & 58.74 & 62.03 & 58.25 & 65.75 \\
        MassiveScenarioClassification (da) & 66.15 & 67.76 & 65.24 & 69.26 \\
        NordicLangClassification & 41.73 & 48.40 & 35.38 & 50.74 \\
        \hline
    \end{tabular}
    \caption{MTEB Danish classification results.}
    \label{tab:appendix/mteb_da_classification}
\end{table}

\begin{table}[h]
    \centering
    \begin{tabular}{l|l|l|l|l}
        \hline
        Dataset & DAP & SONAR & LaBSE & MEXMA \\
        \hline
        Average & 51.58 & 55.59 & 50.76 & 58.08 \\
        MassiveIntentClassification & 55.85 & 59.90 & 57.91 & 64.48 \\
        MassiveScenarioClassification & 62.67 & 65.81 & 64.29 & 68.22 \\
        NoRecClassification & 46.06 & 48.25 & 45.44 & 48.88 \\
        NordicLangClassification & 41.73 & 48.40 & 35.38 & 50.74 \\
        \hline
    \end{tabular}
    \caption{MTEB Norwegian classification results.}
    \label{tab:appendix/mteb_ng_classification}
\end{table}

\begin{table}[h]
    \centering
    \begin{tabular}{l|l|l|l|l}
        \hline
        Dataset & DAP & SONAR & LaBSE & MEXMA \\
        \hline
        Average & 53.03 & 55.09 & 56.00 & 57.09 \\
        AllegroReviews & 31.58 & 29.62 & 34.89 & 31.09 \\
        MassiveIntentClassification (pl) & 58.53 & 65.86 & 59.71 & 66.85 \\
        MassiveScenarioClassification (pl) & 63.05 & 69.99 & 64.58 & 70.20 \\
        PAC & 67.97 & 73.87 & 68.11 & 73.31 \\
        PolEmo2.0-IN & 61.75 & 52.80 & 64.00 & 59.10 \\
        PolEmo2.0-OUT & 35.32 & 38.40 & 44.72 & 42.00 \\
        \hline
    \end{tabular}
    \caption{MTEB Polish classification results.}
    \label{tab:appendix/mteb_pl_classification}
\end{table}

\subsection{Pair Classification}
Pair classification results for English, French and Chinese are reported in \Cref{tab:appendix/mteb_en_pair_classification}, \Cref{tab:appendix/mteb_fr_pair_classification}, and \Cref{tab:appendix/mteb_zh_pair_classification}, respectively.
MEXMA outperforms all other models on average.
\begin{table}[h]
    \centering
    \begin{tabular}{l|l|l|l|l}
        \hline
        Dataset & DAP & SONAR & LaBSE & MEXMA \\
        \hline
        Average & 63.87 & 70.73 & 69.75 & 74.39 \\
        PawsX & 55.30 & 75.05 & 54.07 & 73.18 \\
        SprintDuplicateQuestions & 72.47 & 77.08 & 89.26 & 86.89 \\
        XNLI & 63.83 & 60.06 & 65.92 & 63.10 \\
        \hline
    \end{tabular}
    \caption{MTEB English pair classification results.}
    \label{tab:appendix/mteb_en_pair_classification}
\end{table}

\begin{table}[h]
    \centering
    \begin{tabular}{l|l|l|l|l}
        \hline
        Dataset & DAP & SONAR & LaBSE & MEXMA \\
        \hline
        Average & 73.03 & 77.57 & 73.70 & 78.13 \\
        PawsX (fr) & 55.57  & 71.36 & 54.63 & 71.07 \\
        Opusparcus (fr) & 100.00 & 100.00 & 100.00 & 100.00 \\
        XNLI & 63.52 & 61.34 & 66.48 & 63.32 \\
        \hline
    \end{tabular}
    \caption{MTEB French pair classification results.}
    \label{tab:appendix/mteb_fr_pair_classification}
\end{table}

\begin{table}[h]
    \centering
    \begin{tabular}{l|l|l|l|l}
        \hline
        Dataset & DAP & SONAR & LaBSE & MEXMA \\
        \hline
        Average & 61.12 & 60.80 & 61.95 & 62.12 \\
        PawsX(zh) & 56.20 & 65.35 & 54.26 & 63.68 \\
        Cmnli & 69.29 & 61.86 & 72.67 & 67.45 \\
        Ocnli & 57.86 & 55.18 & 58.91 & 55.23 \\
        \hline
    \end{tabular}
    \caption{MTEB Chinese pair classification results.}
    \label{tab:appendix/mteb_zh_pair_classification}
\end{table}

\subsection{Semantic Textual Similarity (STS)}
Semantic Textual Similarity (STS) results are reported in \Cref{tab:appendix/mteb_en_sts}, \Cref{tab:appendix/mteb_fr_sts}, \Cref{tab:appendix/mteb_pl_sts} and \Cref{tab:appendix/mteb_zh_sts} for English, French, Polish and Chinese, respectively.
LaBSE outperforms MEXMA and the remaining models on STS.
MEXMA and LaBSE outperform SONAR by large margins.
\begin{table}[h]
    \centering
    \begin{tabular}{l|l|l|l|l}
        \hline
        Dataset & DAP & SONAR & LaBSE & MEXMA \\
        \hline
        Average & 67.45 & 67.24 & 70.93 & 70.62 \\
        BIOSSES & 70.51 & 79.11 & 78.70 & 75.97 \\
        SICK-R & 69.18 & 62.94 & 69.99 & 66.00 \\
        STS12 & 64.69 & 65.46 & 65.08 & 67.32 \\
        STS13 & 63.50 & 62.79 & 67.98 & 67.05 \\
        STS14 & 61.49 & 57.54 & 64.03 & 62.73 \\
        STS15 & 75.38 & 74.25 & 76.59 & 75.72 \\
        STS16 & 68.00 & 75.73 & 72.98 & 76.93 \\
        STS17 (en-en) & 77.03 & 79.94 & 79.45 & 80.97 \\
        STS22 (en) & 53.38 & 47.12 & 60.97 & 57.11 \\
        STSBenchmark & 69.39 & 67.39 & 72.25 & 73.53 \\
        STSBenchmarkMultilingualSTS (en) & 69.39 & 67.39 & 72.25 & 73.53 \\
        \hline
    \end{tabular}
    \caption{MTEB English STS results.}
    \label{tab:appendix/mteb_en_sts}
\end{table}

\begin{table}[h]
    \centering
    \begin{tabular}{l|l|l|l|l}
        \hline
        Dataset & DAP & SONAR & LaBSE & MEXMA \\
        \hline
        Average & 45.31 & 42.15 & 47.50 & 51.56 \\
        ATEC & 28.01 & 26.18 & 26.61 & 29.68 \\
        BQ & 40.01 & 37.66 & 42.60 & 44.37 \\
        LCQMC & 54.97 & 50.11 & 52.19 & 61.34 \\
        PAWSX & 12.99 & 32.75 & 10.23 & 27.77 \\
        STS22(zh) & 52.05 & 52.82 & 63.02 & 63.49 \\
        STSB & 63.67 & 50.18 & 68.38 & 65.75 \\
        STSBenchmarkMultilingualSTS (zh) & 65.46 & 45.33 & 69.50 & 68.55 \\
        \hline
    \end{tabular}
    \caption{MTEB Chinese STS results.}
    \label{tab:appendix/mteb_zh_sts}
\end{table}

\begin{table}[h]
    \centering
    \begin{tabular}{l|l|l|l|l}
        \hline
        Dataset & DAP & SONAR & LaBSE & MEXMA \\
        \hline
        Average & 67.74 & 65.60 & 74.33 & 70.10 \\
        SICKFr & 66.84 & 66.1 & 69.94 & 65.94 \\
        STS22 (fr) & 64.44 & 61.72 & 77.95 & 72.19 \\
        STSBenchmarkMultilingualSTS (fr) & 71.92 & 68.99 & 75.1 & 72.17 \\
        \hline
    \end{tabular}
    \caption{MTEB French STS results.}
    \label{tab:appendix/mteb_fr_sts}
\end{table}

\begin{table}[h]
    \centering
    \begin{tabular}{l|l|l|l|l}
        \hline
        Dataset & DAP & SONAR & LaBSE & MEXMA \\
        \hline
        Average                             & 57.06 & 57.17 & 65.82 & 63.67 \\
        CDSC-R                              & 74.12 & 85.77 & 85.53 & 85.95 \\
        SICK-R-PL                           & 60.63 & 62.98 & 65.90 & 64.31 \\
        STS22 (pl)                          & 28.16 & 25.31 & 39.28 & 32.51 \\
        STSBenchmarkMultilingualSTS (pl)    & 65.31 & 54.62 & 72.58 & 71.93 \\
        \hline
    \end{tabular}
    \caption{MTEB Polish STS results.}
    \label{tab:appendix/mteb_pl_sts}
\end{table}

%% file: sections/appendix_token_level_analysis.tex
\section{Token level analysis}\label{section:appendix/token_level_analysis}
\begin{figure}[h]
    \centering
    \includegraphics[width=.6\linewidth]{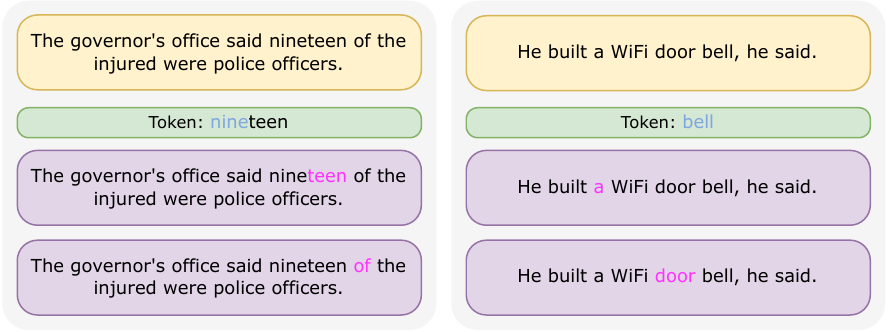}
    \caption{\label{fig:analysis/labse_token_matching_example}Example of LaBSE's token matching. The token in blue is the query token, the tokens in pink are the closest tokens to the query token in the space.}
\end{figure}

\begin{figure}[h]
    \centering
    \includegraphics[width=.6\linewidth]{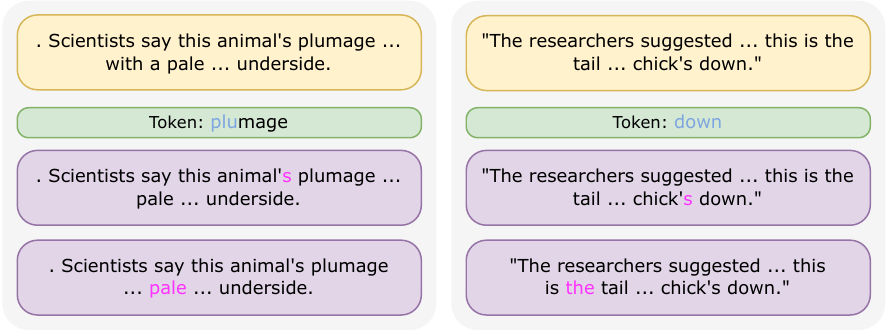}
    \caption{\label{fig:analysis/mexma_no_tok_token_matching_example}Example of MEXMA no token-level grad's token matching. The token in blue is the query token, the tokens in pink are the closest tokens to the query token in the space.}
\end{figure}

\begin{figure}[h ]
    \centering
    \includegraphics[width=.6\linewidth]{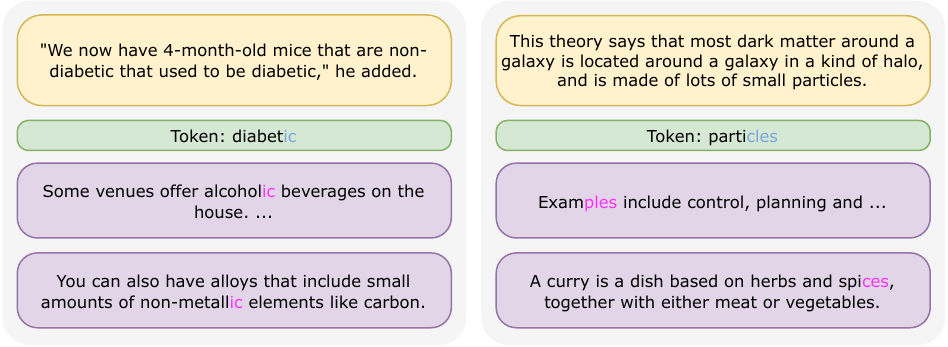}
    \caption{\label{fig:analysis/xlm_roberta_matching_example}Example of XLM-RoBERTa token matching. The token in blue is the query token, the tokens in pink are the closest tokens to the query token in the space.}
\end{figure}

\begin{figure}
    \centering
    \includegraphics[width=\linewidth]{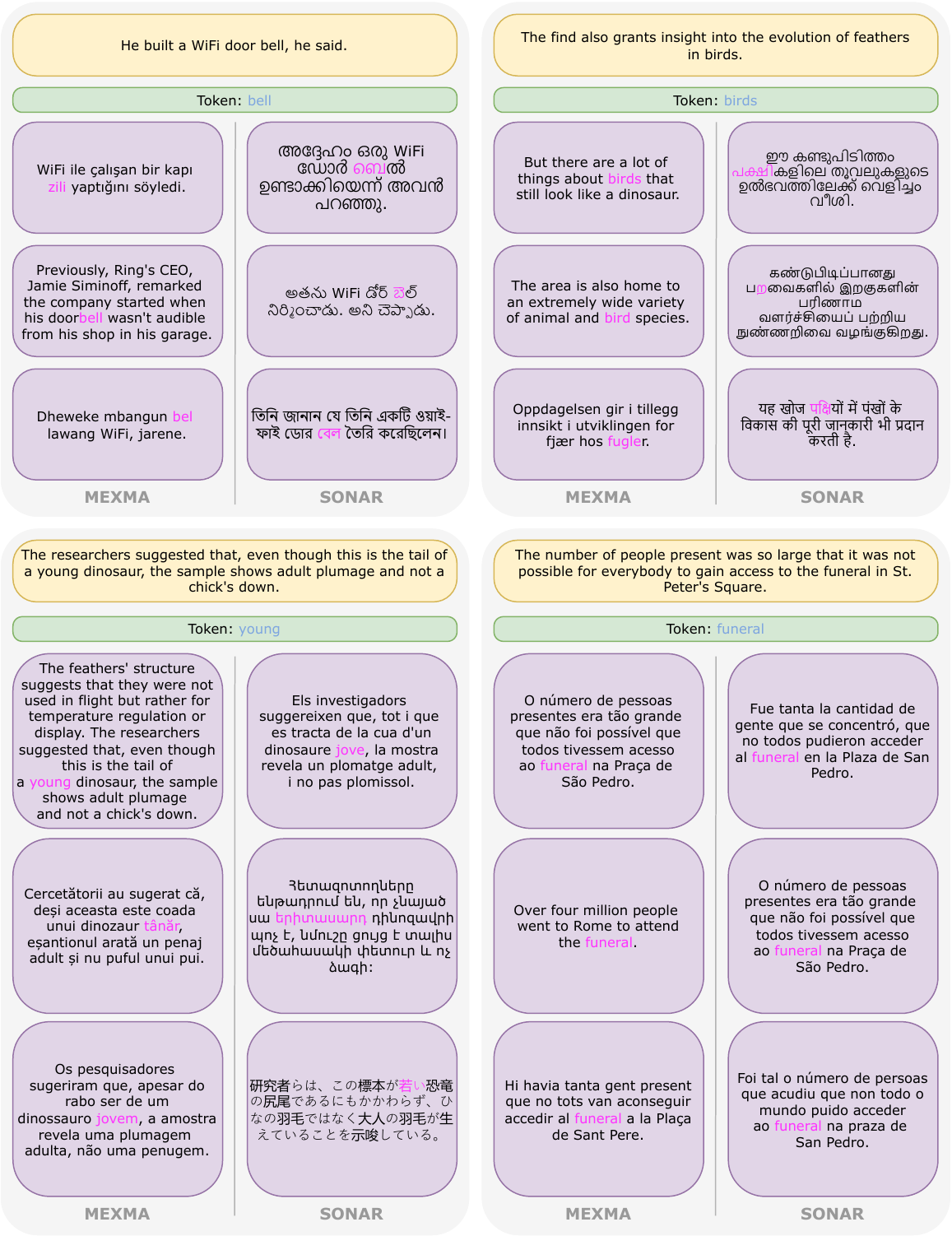}
     \caption{\label{fig:analysis/mexma_sonar_same_lang}Comparison of SONAR and MEXMA token matching. MEXMA displays the ability to match a token in another sentence in the same language. SONAR matches a translated token. The token in blue is the query token, the tokens in pink are the closest tokens to the query token in the space. MEXMA is on the left, SONAR on the right.}
\end{figure}

\begin{figure}
    \centering
    \includegraphics[width=\linewidth]{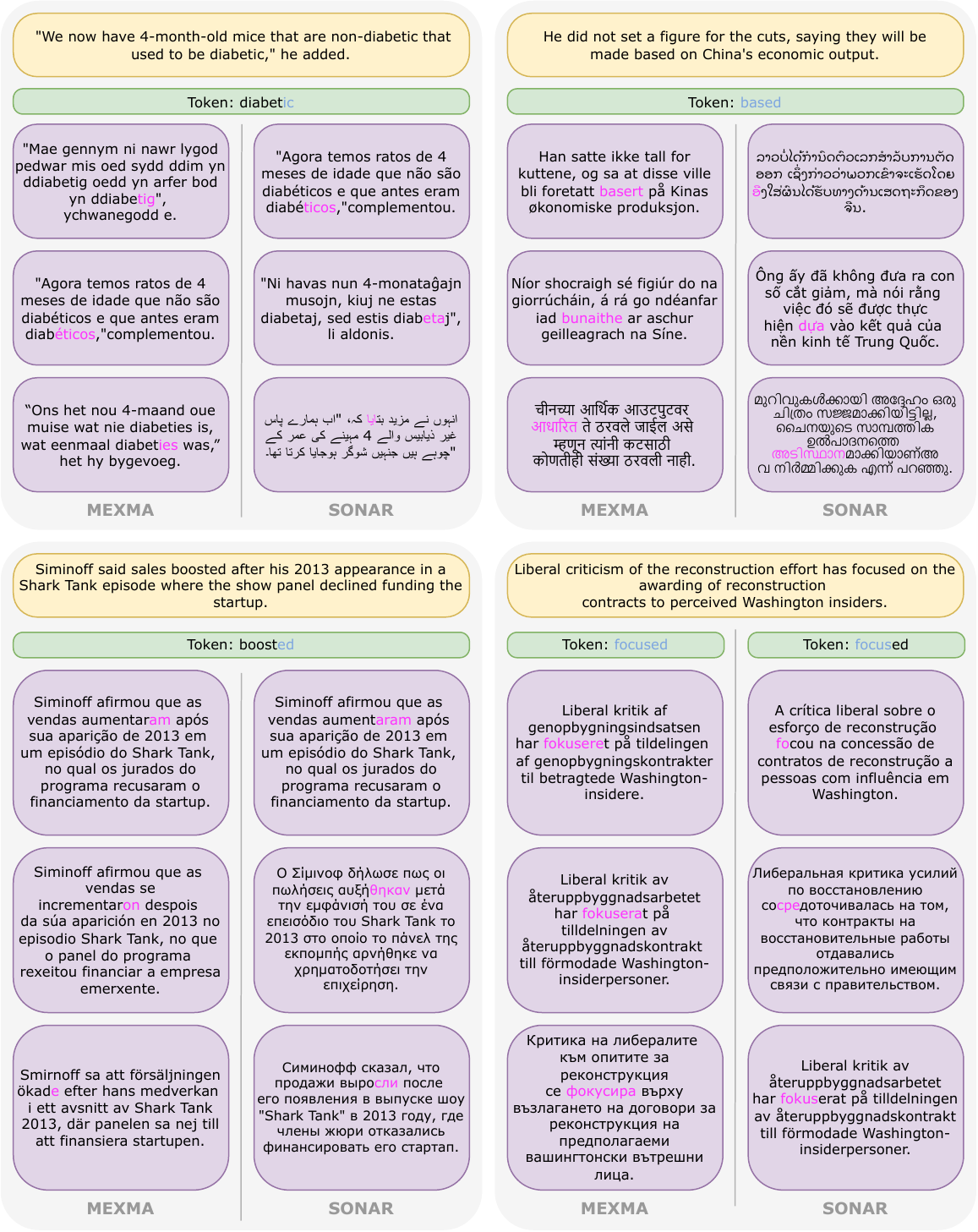}
    \caption{\label{fig:analysis/mexma_sonar_translation}Comparison of SONAR and MEXMA on translated tokens in translations.}
\end{figure}

In this section, we illustrate the behaviour of each model by visualizing the closest tokens in the space.
We observe that MEXMA matches tokens in translations but also different contexts when tokens are used with the same meaning.
This is further broken down in Table \ref{tab:analysis/token_matching_breakdown}, which distinguishes between two types of matches MEXMA does: (1) "same language" matches, where the model identifies the same token used in a different context (monolingual), and (2) "other" matches, where it recognizes translated tokens in a sentence in another language that is not a translation (multilingual).
We observe that SONAR primarily matches tokens across translations, but does not tend to match the same token when it appears in different sentences within the same language.
Examples of MEXMA and SONAR comparisons of matching the same token in other sentences is in \Cref{fig:analysis/mexma_sonar_same_lang}, and both models matching translations in \Cref{fig:analysis/mexma_sonar_translation}.
In both figures, we show the three closest tokens to the selected token, denoted as query on the green box, with the blue text.
The closest tokens are in the purple boxes with the pink text.
Additionally, we show examples of how LaBSE and MEXMA without direct token-level gradients (no-tok MEXMA), match adjacent tokens in the same sentence regularly.
These are shown for LaBSE in \Cref{fig:analysis/labse_token_matching_example}, and for no-tok MEXMA in \Cref{fig:analysis/mexma_no_tok_token_matching_example}.
Lastly, we show how XLM-RoBERTa mostly matches the same tokens in other sentences in the same language, presented in \Cref{fig:analysis/xlm_roberta_matching_example}.
For these last three models, we show the top-2 closest tokens, with the same color scheme as mentioned above.
Each image has two examples for the given model.

%% file: sections/appendix_sentence_analysis.tex
\section{Attention distribution over tokens} \label{appendix:attn_distribution_tokens}
\begin{figure}
    \centering
    \includegraphics[width=\linewidth]{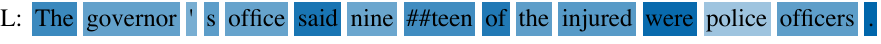} \\
    \includegraphics[width=\linewidth]{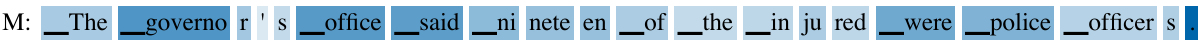} \\
    \hrule
    \includegraphics[width=\linewidth]{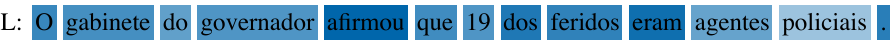} \\
    \includegraphics[width=\linewidth]{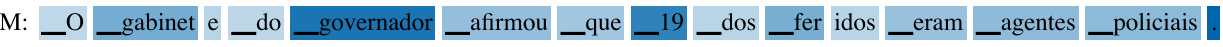} \\
    \caption{Comparison of LaBSE and MEXMA's probabilities distribution over the tokens. In this example, the models had to match the sentence with "19" in Portuguese and English. LaBSE's entries are preceeded with "L:", and MEXMA's with "M:".}
    \label{fig:appendix/attn_distribution_tokens/19_example}
\end{figure}

\begin{figure}
    \centering
    \includegraphics[width=\linewidth]{figures/sentence_probs_analysis/new/new_labse_19_en.pdf} \\
    \includegraphics[width=\linewidth]{figures/sentence_probs_analysis/new/new_mexma_19_en.pdf} \\
    \hrule
    \includegraphics[width=\linewidth]{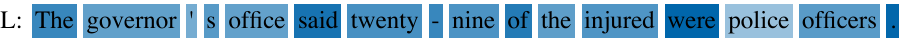} \\
    \includegraphics[width=\linewidth]{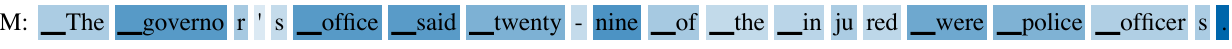} \\
    \caption{Comparison of LaBSE and MEXMA's probabilities distribution over the tokens. In this example, the models had to distinguish the sentence with "19" and "29" in Portuguese and English. LaBSE's entries are preceeded with "L:", and MEXMA's with "M:"}
    \label{fig:appendix/attn_distribution_tokens/29_example}
\end{figure}

\begin{figure}
    \centering
    \includegraphics[width=\linewidth]{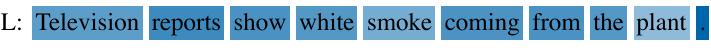} \\
    \includegraphics[width=\linewidth]{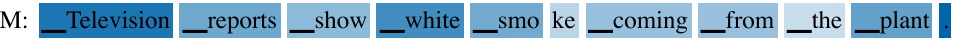} \\
    \hrule
    \includegraphics[width=\linewidth]{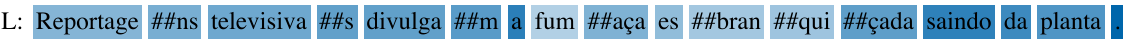} \\
    \includegraphics[width=\linewidth]{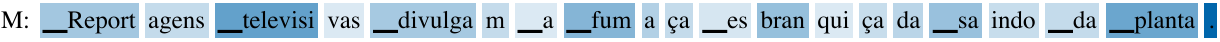} \\
    \caption{Comparison of LaBSE and MEXMA's probabilities distribution over the tokens. In this example, the models had to match the sentence with "white" in Portuguese and English. LaBSE's entries are preceeded with "L:", and MEXMA's with "M:"}
    \label{fig:appendix/attn_distribution_tokens/white_example}
\end{figure}

\begin{figure}
    \centering
    \includegraphics[width=\linewidth]{figures/sentence_probs_analysis/new/new_labse_white_en.pdf} \\
    \includegraphics[width=\linewidth]{figures/sentence_probs_analysis/new/new_mexma_white_en.pdf} \\
    \hrule
    \includegraphics[width=\linewidth]{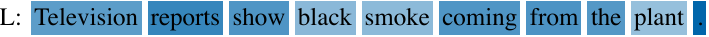} \\
    \includegraphics[width=\linewidth]{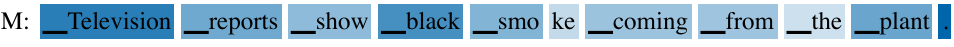} \\
    \caption{Comparison of LaBSE and MEXMA's probabilities distribution over the tokens. In this example, the models had to distinguish the sentence with "white" and "black" in Portuguese and English. LaBSE's entries are preceeded with "L:", and MEXMA's with "M:"}
    \label{fig:appendix/attn_distribution_tokens/white_black_example}
\end{figure}

In this section, we provide some examples of MEXMA and LaBSE's attention probabilities given by the CLS token to the word tokens.
The examples are provided in Figures \ref{fig:appendix/attn_distribution_tokens/19_example}, \ref{fig:appendix/attn_distribution_tokens/29_example}, \ref{fig:appendix/attn_distribution_tokens/white_example} and \ref{fig:appendix/attn_distribution_tokens/white_black_example}.
Across all figures, it is possible to see that LaBSE tends to be more uniform across all tokens, while MEXMA tends to focus more attention on a smaller subset of the tokens.
All examples are taken from the FLORES200 test set with the xsim++ extension, where some words in the original sentences are replaced, and the models have to be able to still match the correct translation, and not a sentence with a small change.
From \Cref{fig:appendix/attn_distribution_tokens/19_example} to \Cref{fig:appendix/attn_distribution_tokens/29_example} "nineteen" is replaced with "twenty nine". 
From \Cref{fig:appendix/attn_distribution_tokens/white_example} to \Cref{fig:appendix/attn_distribution_tokens/white_black_example} the word "white" is replaced with "black".

\Cref{fig:appendix/attn_distribution_tokens/19_example} shows the attention placed by MEXMA and LaBSE on the same sentence in English and Portuguese.
It is possible to see that MEXMA in Portuguese places most of the attention in two tokens, "governador" and "19", where the token in "19" is very relevant here since it is the one needed to distinguish the examples in xsim++.
LaBSE seems to have many tokens with a lot of attention, and does not have "19" as one of the tokens with the most attention.

In \Cref{fig:appendix/attn_distribution_tokens/29_example}, we have the English example with nineteen (as previously shown in \Cref{fig:appendix/attn_distribution_tokens/19_example}) compared to the same sentence with nineteen replaced by twenty-nine.
Interestingly, LaBSE places more attention on the "\#\#teen" token than the "nine" token, but similar attention to the "twenty", "-" and "nine" tokens.
MEXMA places similar attention in all nineteen tokens, and in twenty nine it places a small amount of attention on the irrelevant "-", with a higher degree of attention in "nine" and a smaller amount of attention in "twenty".
MEXMA also seems to do a good job ignoring irrelevant tokens like "of", while LaBSE places a lot of attention in it.

\Cref{fig:appendix/attn_distribution_tokens/white_example} has the same sentence in English and Portuguese, where, in xsim++ the models need to be able to match the color "white" instead of other colors.
It is possible to see that, for LaBSE, white is not one of the most relevant tokens in English, but for MEXMA it is, along with "television".
In Portuguese the behavior is similar, the token "bran" in "esbranquiçada" has a large degree of attention from MEXMA, while for LaBSE is it not a token with a lot of attention, and "çada" which is a token that does not convey the idea of white, is the one with the most attention out of the 4 tokens of the word, for LaBSE.
In Portuguese it is also noticeable that MEXMA gives a small amount of attention to most of the less relevant tokens, while LaBSE seems to have a lot more tokens with a high degree of attention.

\Cref{fig:appendix/attn_distribution_tokens/white_black_example} shows the same English sentence as \Cref{fig:appendix/attn_distribution_tokens/white_example}, with the word white replaced with the word black.
Interestingly, MEXMA's attention remains the same with black and white, while for LaBSE the token "black" seems to get less attention than the token "white".
The remaining tokens get similar attention in both models.

\begin{figure}
    \centering
    \includegraphics[width=0.48\linewidth]{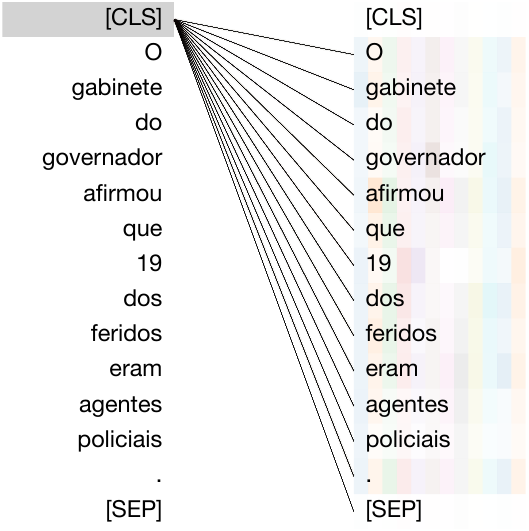} \hfill
    \includegraphics[width=0.48\linewidth]{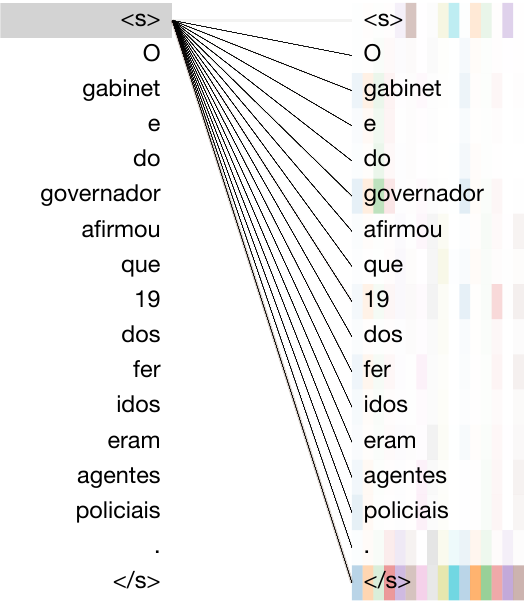}
    \caption{Attention distribution of MEXMA and LaBSE across all heads, and all tokens. On the left is LaBSE, on the right is MEXMA. MEXMA uses the EOS token as an attention dump, and has a more skewed distribution, while LaBSE has a more uniform distribution.}
    \label{fig:appendix/attn_distribution_tokens/mexma_vs_labse_bertviz}
\end{figure}

Additionally, \Cref{fig:appendix/attn_distribution_tokens/mexma_vs_labse_bertviz}, provides a comparison for MEXMA and LaBSE with the probabilities of all heads, and all tokens, using BertViz \citep{vig-2019-multiscale}.
It is possible to see that MEXMA places a lot of attention on the EOS token, $<$/s$>$, which is used as an attention dump, i.e. an irrelevant token that receives a very large attention probability, a common phenomena in transformers, as explored in \citet{xiao2024efficientstreaminglanguagemodels, darcet2024visiontransformersneedregisters, sun2024massiveactivationslargelanguage}.
This happens frequently with MEXMA.
It is, again, possible to see the difference in uniformity for MEXMA and LaBSE, with LaBSE having a more uniform attention in the figure.
If we remove the BOS and EOS tokens from the entropy computation, we now get an entropy of $\approx$ 3.5 and $\approx$ 3 for LaBSE and MEXMA, respectively. 
MEXMA's entropy increases, while LaBSE stays mostly similar, which shows that MEXMA indeed frequently uses the EOS token as a dump.
However, MEXMA still has a lower entropy and a more skewed distribution over its word tokens, with or without BOS and EOS, as shown by the lower entropy and the Figures \ref{fig:appendix/attn_distribution_tokens/19_example}-\ref{fig:appendix/attn_distribution_tokens/white_black_example}.

%% file: sections/appendix_other_architectures.tex
\section{Baseline architectures}
\begin{figure*}[h]
    \centering
    \begin{subfigure}[t]{0.49\textwidth}
        \centering
        \includegraphics{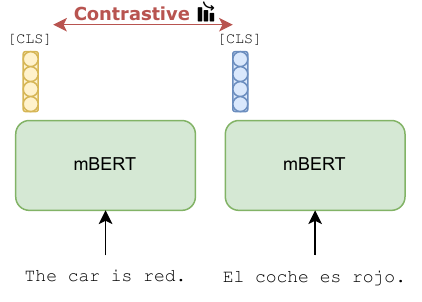}
        \caption{LaBSE's architecture.}
        \label{fig:appendix/labse-architecture}
    \end{subfigure}%
    ~ 
    \begin{subfigure}[t]{0.49\textwidth}
        \centering
        \includegraphics{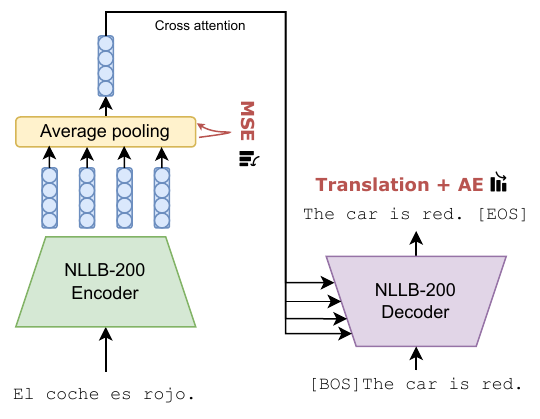}
        \caption{SONAR's architecture.}
        \label{fig:appendix/sonar-architecture}
    \end{subfigure}
    \begin{subfigure}[t]{0.55\textwidth}
        \centering
        \includegraphics{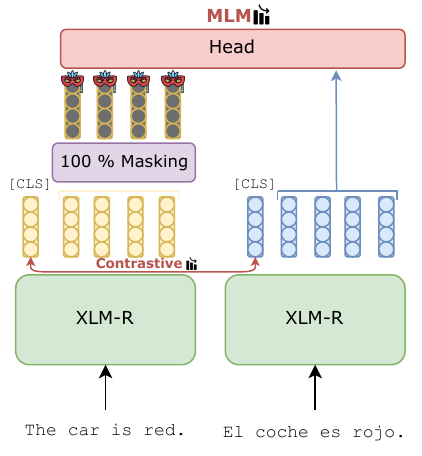}
        \caption{DAP's architecture.}
        \label{fig:appendix/dap-architecture}
    \end{subfigure}%
    ~ 
    \begin{subfigure}[t]{0.6\textwidth}
        \centering
        \includegraphics{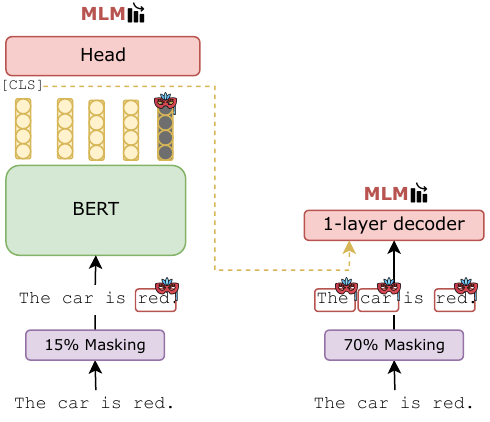}
        \caption{RetroMAE's architecture.}
        \label{fig:appendix/retromae-architecture}
    \end{subfigure}
    \caption{Architecture of the baselines.}
\end{figure*}

We report SONAR, LaBSE's, DAP's and RetroMAE's architectures in Figures \ref{fig:appendix/sonar-architecture}, \ref{fig:appendix/labse-architecture}, \ref{fig:appendix/dap-architecture} and \ref{fig:appendix/retromae-architecture}, respectively for easier comparison.
LaBSE employs a slightly modified contrastive loss, to increase separation, and SONAR is based on translation.
DAP uses token-level objectives, but it does not leverage them to update the sentence representation.
RetroMAE uses the sentence in the heavy unmasking, but that unmasking does not update the tokens outputted by the encoder, it is monolingual, and the sentence representation does not come from an unmasked input.
MEXMA is based on cross unmasking and has direct token level gradients updating its internal representations.

